\documentclass[11pt,letterpaper]{article}
\usepackage{arxiv}
\usepackage[linesnumbered,ruled,vlined]{algorithm2e}
\usepackage[pdftex]{graphicx}
\usepackage{graphicx}
\usepackage{rotate}
\usepackage{wrapfig}
\usepackage{caption}
\usepackage[T1]{fontenc}
\usepackage[english]{babel}
\usepackage{times}
\usepackage{amsmath,amstext,amsbsy,amsopn,amsthm,amsfonts,amssymb}
\usepackage{mathtools}
\usepackage{url}
\usepackage{tabularx,booktabs}
\newcolumntype{C}{>{\centering\arraybackslash}X}
\usepackage{booktabs,array,colortbl}
\usepackage{lastpage}
\usepackage{multirow}
\usepackage{balance}
\usepackage{placeins}
\usepackage{array}
\usepackage{blindtext}
\usepackage{subfigure}
\usepackage{caption}
\usepackage{capt-of}
\usepackage{booktabs}% for better rules in the table
\usepackage[small]{titlesec}
\usepackage{paralist}
\usepackage{verbatim} % For comment
\usepackage[usenames,dvipsnames,svgnames]{xcolor}
\usepackage{algorithmic}
\usepackage{fancyhdr}
\usepackage{wrapfig}
\providecommand{\keywords}[1]{\textbf{\textit{Keywords---}} #1}
\begin{document}

\title{Accu-Help: A Machine Learning based Smart Healthcare Framework for Accurate Detection of Obsessive Compulsive Disorder}

%\title{iGLU: An Edge-Device for Non-Invasive Precision Blood Glucose Monitoring in Smart Healthcare}

%\title{iGLU: A NIR based Machine-Learning Integrated Edge-Device for Non-Invasive Precision Blood Glucose Monitoring in Smart Healthcare}
%\author[4]{\fnm{Susanta K.} \sur{Padhy}}\email{psych\_susanta@aiimsbhubaneswar.edu.in}

\author{
\begin{tabular}{cccc}
	\centering
Kabita Patel$^1$ & Ajaya K. Tripathy$^{2}$\thanks{corresponding author} & Laxmi N. Padhy$^3$ & Sujita K. Kar$^4$ \\
& Susanta K. Padhy$^5$ & Saraju P. Mohanty$^6$ &  
\end{tabular}\\
$^{1,2}$Dept. of CS, Gangadhar Meher University, India.\\
$^3$Dept. of CSE, Konark Institute of Science and Technology, India.\\
$^4$Dept. of Psychiatry, King George's Medical University,  India.\\
$^5$ Department of Psychiatry, All India Institute of Medical Sciences, Bhubaneswar, India.\\
$^6$Dept. of Computer Sci. and Eng., University of North Texas, USA\\
Email: $^1$kabitapatelgmu@gmail.com, $^2$ajayatripathy1@gmail.com, $^3$lnpadhy2020@gmail.com,\\
$^4$drsujita@gmail.com, $^5$psych\_susanta@aiimsbhubaneswar.edu.in, and $^6$saraju.mohanty@unt.edu
%Prateek~Jain,~\IEEEmembership{Graduate Student Member,~IEEE};
%Amit M.~Joshi,~\IEEEmembership{Member,~IEEE};\\
%and Saraju P. Mohanty,~\IEEEmembership{Senior Member,~IEEE}% <-this % stops a space
%\thanks {P. Jain is with the Department of Electronics and Communications Engineering (ECE), Malaviya National Institute of Technology, Jaipur, Email: prtk.ieju@gmail.com.} %\protect\\
%\thanks{A. M.~Joshi is with the Department of Electronics and Communications Engineering (ECE), Malaviya National Institute of Technology, Jaipur, E-mail: amjoshi.ece@mnit.ac.in.} %\protect\\
%\thanks{S. P. Mohanty is with the Department of Computer Science and Engineering, University of North Texas, E-mail: saraju.mohanty@unt.edu.}%\protect\\
}

\maketitle

\cfoot{Page -- \thepage-of-\pageref{LastPage}}

\begin{abstract}
	In recent years the importance of Smart Healthcare can’t be overstated. The current work proposed to expand the state-of-art of smart healthcare in integrating solutions for Obsessive Compulsive Disorder (OCD). Identification of OCD from oxidative stress biomarkers (OSBs) using machine learning is an important development in the study of OCD. However, this process involves the collection of OCD class labels from hospitals, collection of corresponding OSBs from biochemical laboratories, integrated and labeled dataset creation, use of suitable machine learning algorithm for designing OCD prediction model, and making these prediction models available for different biochemical laboratories for OCD prediction for unlabeled OSBs. Further, from time to time, with significant growth in the volume of the dataset with labeled samples, redesigning the prediction model is required for further use. The whole process requires distributed data collection, data integration, coordination between the hospital and biochemical laboratory, dynamic machine learning OCD prediction mode design using a suitable machine learning algorithm, and making the machine learning model available for the biochemical laboratories. Keeping all these things in mind, Accu-Help a fully automated, smart, and accurate OCD detection conceptual model is proposed to help the biochemical laboratories for efficient detection of OCD from OSBs. OSBs are classified into three classes: Healthy Individual (HI), OCD Affected Individual (OAI), and Genetically Affected Individual (GAI). The main component of this proposed framework is the machine learning OCD prediction model design. In this Accu-Help, a neural network-based approach is presented with an OCD prediction accuracy of $86\pm2\%$
\end{abstract}


%\IEEEpeerreviewmaketitle

%%%%%%%%%%%%%%%%%%%%%%%%%%%%%%%%%%%%%%%%%%%%%%%%%
\section{Introduction}
\label{sec:introduction}
With the improvement of machine learning, the internet-of-things, machine learning techniques, and cyber-physical system technology, there is a wide scope of enhancing the quality of healthcare systems. Recently, several researchers have demonstrated their enthusiasm in designing and developing smart healthcare systems to address different issues of the healthcare system. For example, to automate the seizure detection from EEG the authors in \cite{olokodana2021ezcap} have proposed a  IoT-based System using Machine Learning. 
A smart healthcare system is proposed in \cite{rachakonda2022bactmobile} to detect Blood Alcohol Concentration using machine learning. 
 A fuzzy neural network-based Internet-of-Things system is presented in \cite{kumar2018cloud} for disease prediction. A smart healthcare system is proposed in \cite{amin2019cognitive} to detect and monitor diseases to provide real-time support to patients. Very less focus are given in the literature in  designing H-CPS for mental illnesses like Anxiety, Obsession, compulsion, or both   Obsession and compulsion.

Obsessive-compulsive disorder (OCD) is one of the classes of anxiety illness \cite{vos2017global}. Individuals with OCD suffer from obsession and compulsion. Obsessions are unpleasant and undesired thoughts or feelings that come into mind. Generally, obsession makes an individual very uncomfortable, anxious, and fearful. For the sake of correction, individuals with obsession carry out repeated activities called compulsion. For example, cleaning hands repeatedly to overcome the thought of contamination.

One of the effective treatments for OCD is Cognitive-behavioral therapy (CBT). However, OCD is detected from behavioral symptoms analysis, but the OCD behavior is observable at a letter stage. Earlier detection can help the patient for quick recovery using CBT treatment. However, in general, OCD-affected persons have less trust in the detection process through symptom analysis in the preliminary stage. As long as this OCD problem is not harming their day-to-day activity, they are not accepting the disease and abstaining from treatment. Therefore, the treatment starts at a later stage and the recovery period becomes longer. In such a situation, OCD detection using an intelligent machine may create trust in the diagnosis process of OCD at an early stage. 

\begin{figure}[tbh]
	\centering
	\includegraphics[width=0.8\textwidth, trim=6.5cm 9cm 6.5cm 8.2cm,clip]{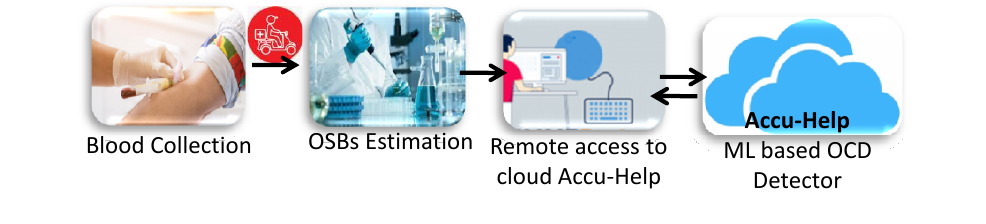}
	\caption{Accu-Help: A Healthcare Cyber-Physical System (H-CPS) for OCD Detection }
	\label{fig:concept1}
\end{figure}

A Healthcare Cyber-Physical System (H-CPS) is proposed called Accu-Help. Accu-Help is a machine learning-based smart healthcare framework for the accurate detection of OCD. H-CPS is an information and communication technology-based infrastructure in which the healthcare process is supervised and controlled using smart systems.  The high-level conceptual picture of the proposed H-CPS is presented in Figure \ref{fig:concept1}. The responsibility of the proposed H-CPS includes the collection of OCD class labels from hospitals, collection of corresponding OSBs from biochemical laboratories, integrated database creation, use of suitable machine learning algorithm for designing OCD prediction model, and making these prediction models available for different biochemical laboratories for OCD prediction for unlabeled OSBs. Further, from time to time, with significant growth in the size of the database with labeled samples, redesigning the prediction model is required for further use.  The whole process requires distributed data collection, data integration, coordination between the hospital and biochemical laboratory, dynamic machine learning OCD prediction mode design using a suitable machine learning algorithm, and making the machine learning model available for the users.  
The core part of the proposed H-CPS is a machine learning model which is capable of providing intelligence to the proposed H-CPS for the detection of OCD. Accu-Help can collect OSBs and OCD class labels from different hospitals and different biochemical laboratories located in different geographical locations. OSBs along with the class labels are used for ML model design for OCD classification and made available online. Further, Accu-Help can be used by biochemical laboratories to detect OCD class labels of unknown OSBs samples.

Accu-Help uses a hyperparameter-optimized neural network for OCD identification or classification using oxidative stress biomarkers.  
The effectiveness of the proposed mechanism is compared with the effectiveness of some of the popular classification approaches. 
Experimental outcomes reveal  that the proposed mechanism is superior in comparison with others.    

The rest of the article is organized as follows. 
The state of the art of the problem at hand is summarized in Section \ref{sec:soda}. 
The novelty of the article is presented in Section  \ref{sec:novelContrubition}.  
In Section \ref{sec:cps} a cyber-physical system is proposed to handle the whole system of OCD detection.  Various popular machine learning methods are used for OCD detection in Section \ref{sec:classification}. 
Section \ref{sec:honn} presents the proposed hyperparameter-optimized neural network for OCD classification through oxidative stress biomarkers. 
The dataset description, experimental result analysis, and comparative studies are performed in Section \ref{sec:exp}. Section \ref{sec:con}  gives the future work direction and concludes the article.

\section{Related Prior Works}
\label{sec:soda}
\subsection{Smart Healthcare Systems}
With the advancement of IoT and smart technologies, there has been an attempt of incorporating healthcare system to automatically diagnosing, managing different expects of diseases \cite{nouman2021recent,jain2019iglu, maji2021ikardo}. However, the proposed healthcare system mainly focuses on OCD and the diagnose of OCD from oxidative stress biomarkers. There are several smart healthcare systems proposed in the literature. The research work in \cite{catarinucci2015iot} proposes a model to automate the process of  automatic monitoring of patients, and biomedical devices. 
To automate the seizure detection from EEG the authors in \cite{olokodana2021ezcap} have proposed a  IoT-based System using Machine Learning. 
A smart healthcare system is proposed in \cite{rachakonda2022bactmobile} to detect Blood Alcohol Concentration using machine learning. In \cite{tripathy2020easyband} the authors have proposed a healthcare system to handle the mobility of individuals during a pandamic. Even though several smart healthcare systems are proposed in the literature, as per the knowledge of the authors none os the approaches has proposed smart healthcare system for OCD detection process.

\subsection{Related Prior Research}
With the growth of OCD cases globally, there has been significant advancements in the research to analyze different aspects of OCD such as detection, observing treatment effectiveness,  and severity prediction. Machine learning approaches were also well explored to address different problems of it. Most of these approaches analyze different aspects of OCD using the biomarkers such as EEG, MRI, fMRI, and DTI.  Several studies focus on the diagnosis of OCD patients by studying neuropsychological biomarkers, genetic biomarkers, MRI biomarkers \cite{mas2016integrating}, fMRI biomarkers, DTI and EEG biomarkers \cite{aydin2015classification,erguzel2015hybrid}, and EEG biomarkers \cite{desarkar2007high}. Artificial intelligence approaches have been used in the detection of OCD treatment effectiveness in \cite{salomoni2009artificial}, OCD severity forecast \cite{hoexter2013predicting} and OCD severity reduction forecast model has been designed in \cite{askland2015prediction,lenhard2018prediction}. In general, the EEG analysis is performed by power and source analysis. EEG channel one and four analyses have been performed for OCD detection in \cite{aydin2015classification}. An intracortical EEG signal is analyzed in \cite{kopvrivova2011eeg} and observed hyper-activation for OCD individuals. MRI and fMRI have been analyzed in many studies to diagnose OCD \cite{weygandt2012fmri,piaggi2014singular, rangaprakash2013phase}. 
However, the collection of such biomarkers involves high-end machines which may not be available in most places. In absence of biological markers such as EEG, MRI, fMRI, and DTI, an alternative mechanism should be adapted. 
The OCD research literature on neuroimaging biomarkers  and their limitations are summarized in Table \ref{tab:neuroimaging}.

\begin{table}[]
	\centering
	\caption{OCD research literature on neuroimaging biomarkers  and their limitations }
	\label{tab:neuroimaging}
	\begin{tabular}{|p{0.12\textwidth}|p{0.45\textwidth}|p{0.35\textwidth}|}
		\hline
		\textbf{Research} &	\textbf{Analysis and results} &	\textbf{Limitations} \\ 
		& & 		\\ \hline
		
		OCD severity detection \cite{mas2016integrating} &	Data Analyzed: MRI, DTI, and neuropsychological data.
		The technique used: Applied machine learning for OCD severity detection
		Result:  Detection Ability of $90\%$  in the training set and $70\%$ in the testing set. &	Collection of such biomarkers requires high-end machines near the patient. Without biological markers such as EEG, MRI, fMRI, and DTI, an alternative mechanism should be adopted.  \\ \hline

		Classification of OCD \cite{aydin2015classification} &
		Data Analyzed: EEG data and hemispheric dependency data.
		The technique used: Support vector machine (SVM) classifiers.
		Result: Achieved an OCD classification accuracy of $85 \pm 5.2\%$. &	This approach does not able to classify GAI. Further, in the absence of biological markers such as EEG, MRI, fMRI, and DTI, an alternative mechanism should be adapted. 	  \\ \hline
		
		Classify trichotillomania and OCD \cite{erguzel2015hybrid} &
		Data Analyzed: EEG biomarkers.
		The technique used: SVM with ant colony optimization. 
		Result: Achieved a classification accuracy of $81.04\%$. &	This approach is not able to address the OCD classification problem and is not able to identify the GAI group.  \\ \hline
		
		EEG source analysis in OCD \cite{desarkar2007high,kopvrivova2011eeg} &	Data Analyzed: EEG biomarkers.
		The technique used: Compared resting state using standardized low-resolution electromagnetic tomography.
		Result: Observed that there is a medial frontal hyperactivation in OCD. &	This approach is  not able to classify GAI. Further, an alternative mechanism should be adapted in the absence of biological markers such as EEG, MRI, fMRI, and DTI. 		  \\ \hline
		
		Prediction of OCD treatment response \cite{salomoni2009artificial} &
		Data Analyzed: Symptoms dimension, neuropsychologic act, and epidemiologic parameters.
		The technique used: Multilayer perceptrons.
		Result: $93.3\%$ of correct classification of cases achieved. &	This approach is not able to address the OCD classification problem. Does not able to identify GAI group. 		  \\ \hline
		Predicting OCD severity \cite{hoexter2013predicting} &
		Data Analyzed: MRI data.
		The technique used: support vector regression.
		Result: Concluded that Support Vector Regression can predict OCD symptom severity. &	This approach is not able to address the OCD classification problem. Does not able to identify GAI group. 		  \\ \hline
		
		fMRI pattern recognition in OCD \cite{weygandt2012fmri} &	Data Analyzed: fMRI data.
		The technique used: Multivariate pattern classification techniques.
		Result: Neurobiological markers provide reliable diagnostic information about OCD. &	This approach is not able to classify GAI. Further, an alternative mechanism should be adapted in the absence of biological markers such as EEG, MRI, fMRI, and DTI. 		 \\ \hline
	
	\end{tabular}
	
\end{table}

 A recent study suggests the act of oxidative stress biomarkers in OCD and these biomarkers can be measured from blood samples \cite{shrivastava2017study}.  This study proposes an approach to segregate the three classes (HI, GAI, and OAI) by analyzing oxidative stress biomarkers. The oxidative stress biomarkers considered for the study are superoxide dismutase (SD), Glutathione Peroxidase (GP), Catalase (CAT), Malondialdehyde (MAL), and serum cortisol (SC).
Several studies were performed in the literature on the analysis of oxidative stress biomarkers individually. In some studies \cite{shrivastava2017study,kuloglu2002antioxidant} COR is found to be normal whereas in some other studies \cite{kluge2007increased,gehris1990urinary} it is found to be high in the case of OCD individuals. As per \cite{shrivastava2017study,kuloglu2002antioxidant,ersan2006examination,shohag2012serum}, CAT, GPX, and SOD are lower and MDA is higher in OCD whereas levels of CAT, GPX, and SOD were higher in OCD individuals in other studies \cite{kuloglu2002antioxidant, behl2010relationship}. In \cite{rangaprakash2013phase} the authors have made a careful study of oxidative stress biomarkers and it is observed that these markers act as a major role in OCD individuals \cite{rangaprakash2013phase}. In \cite{rangaprakash2013phase} it is also observed that the population mean and standard deviation are significantly different in the case of OCD individuals. However, it is not possible to predict OCD through the statistical analysis of each marker. 
The aim of this research is to design an intelligent predictive method to predict the existence of OCD in an individual through oxidative stress biomarkers. Thus, it is essential not only to develop a mechanism to detect OCD without EEG, MRI, fMRI, and DTI but also to achieve high detection accuracy. The OCD research literature on oxidative stress biomarkers  and their limitations are summarized in Table \ref{tab:osb}.

\begin{table}[]
	\centering
	\caption{OCD research literature on oxidative stress biomarkers  and their limitations }
	\label{tab:osb}
	\begin{tabular}{|p{0.2\textwidth}|p{0.36\textwidth}|p{0.36\textwidth}|}
		\hline
		\textbf{Research} &	\textbf{Findings} &	\textbf{Limitations} \\ 
	 & &	\\ \hline
	Urinary free cortisol (UFC) Cortisol level analysis in OCD \cite{gehris1990urinary} &
	UFC of both the groups was compared and the OAI group had significantly higher UFC levels than the HI group. &	This study analyzed the cortisol level but had not come up with a threshold as OCD detector.  \\ \hline
	
	MAL, SD, GP, and CAT  levels in patients with OCD \cite{kuloglu2002antioxidant} &
	Higher MAL, SD, GP, and CAT activity but the differences were not big.
However, it is observed that OCD is linked with free radicals. &	This study analyzed oxidative stress biomarkers but had not come up with a threshold as an OCD detector. \\ \hline
	
	Analysis of free radical metabolism and antioxidants in OCD \cite{ersan2006examination} &
	A higher level of MAL was observed in the OCD group. &	This study analyzed MAL but had not come up with a threshold as an OCD detector. \\ \hline
	
	Analysis of oxidative stress of OCD patients \cite{behl2010relationship} &	Oxidative stress biomarkers imbalance was observed in the OCD group. &	This study analyzed oxidative/ antioxidative status but had not come up with a threshold as an OCD detector. \\ \hline
	
	Oxidative stress biomarkers analysis in three groups (HI, GAI, and OAI) \cite{shrivastava2017study} &	Levels of CAT, SD and GP in all three groups are significantly different. &	This study analyzed oxidative stress in all three groups but had not come up with a mechanism to segregate these three groups. \\ \hline
	
	Accu-Help: A Machine Learning-based smart healthcare framework for accurate detection of OCD class  (HI, GAI, or OAI).  & Hyperparameter optimized Neural networks have achieved a prediction accuracy of  $86 \pm 1$\%.  &  This approach is successful in applying Neural Networks on OSBs to predict OCD class. However, the training process is computationally costly.   \\ \hline
	
	\end{tabular}
\end{table}

\section{Novel Contributions of the Article}
\label{sec:novelContrubition}
Distinguishing the healthy controls from the OCD-affected individuals is considered as the OCD detection process.  The common method of OCD detection is performed by symptom analysis. However, this method of symptom analysis works at the latter stage of the disease. The symptoms are not significantly visible in the early stage, as a result, early detection is not possible. Further, in the case of an early or moderate stage of OCD, the symptoms are mild and in general, the patient doesn’t trust the detection through symptoms. However, laboratory detection of OCD may create trust in the mind of the patient and as a result, they may accept the treatments in the early stages.

Several studies have been performed to detect OCD by applying machine learning to biomarkers such as MRI, fMRI, EEG, and DTI. However, the collection of such biomarkers involves high-end machines which may not be available in most places.A recent study suggests a link between oxidative stress biomarkers in OCD and these biomarkers can be measured from blood samples \cite{kar2016empirical}.   Further, the recent study performed in \cite{shrivastava2017study} suggests the existence of a significantly distinguished pattern in the biomarkers of the first-degree relatives of OCD-affected individuals. This study further suggests the identification or segregation of the three groups (HI, GAI, and OAI) has greater importance in the study of OCD. 

\subsection{Research Questions}
\label{sec:issues}
The primary motive of the research is to propose a Healthcare Cyber-Physical System (Accu-Help) to automate the OCD detection process to improve the idea of "Smart-Healthcare". Using Accu-Help the issues resolved are:

\begin{itemize}
	\item Several approaches are proposed for OCD detection using machine learning by analyzing neuroimaging biomarkers. A collection of such biomarkers involve high-end equipment and the individual needs to be physically present near the equipment for data collection. However, such equipment is not available in many places and the collection of such markers becomes a challenge. As a result, these OCD detections
	approaches become less effective.
	
	\item Several studies found the role of oxidative stress in OCD. Most such studies are limited to only the statistical comparative studies of the oxidative stress biomarkers (OSBs) between the HI group and the OAI group. However, none of these approaches proposed any mechanism to detect OCD just by taking OSBs as input.
	
	\item Some of the studies found genetic linkage to OCD. Therefore, it is also helpful to identify individuals which are genetically linked to OCD.
	
	\item Less importance is given to designing a machine learning prediction model to detect the OCD class (HI, GAI, or OAI) from given OSBs.
	
	\item It is also essential to design an automated distributed system to collect labeled OSBs from hospitals and biochemical laboratories, design a machine learning prediction model, and make it available for future use by
	biochemical laboratories for OCD detection by just giving the OSBs of an individual as input. 
\end{itemize}

\subsection{Proposed Solution}
\begin{figure*}[tbh]
	\centering
	\includegraphics[width=\textwidth]{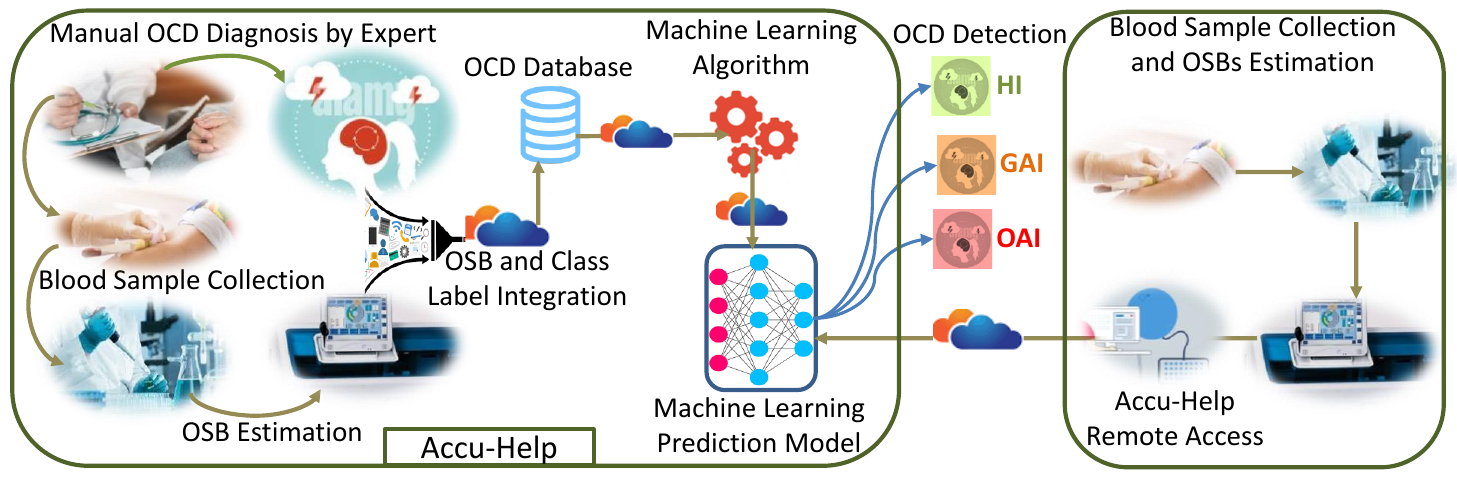}
	\caption{Structure of the proposed Accu-Help Cyber-Physical System (H-CPS) for OCD Detection }
\label{fig:concept}
\end{figure*}

Considering the issues discussed in Section \ref{sec:issues}, Accu-Help a H-CPS is proposed as presented in Figure \ref{fig:concept}. Oxidative stress biomarkers such as SD, GP, CAT, MAL, and SC are estimated from the blood samples of an individual. These biomarkers are passed to a machine learning prediction model designed by the Accu-Help environment to identify the class (one among HI, GAI, and  OAI) by analyzing oxidative stress biomarkers. Accu-Help collects labeled OSBs from hospitals and biochemical laboratories. Accu-Help used a neural network-based classification approach to design the OCD prediction model. In this study, OCD detection signifies the identification of the class (one of the three classes: HI, GAI, and  OAI) from someone's oxidative stress biomarkers. 

\subsection{Research Objective}
The concept behind Accu-Help is designed by taking into account the process of  OCD detection reachability of OCD-related individuals and its ancillary impact on the community. The aims that Accu-Help tries to  address are:

\begin{description}
	\item[OCD Individuals Health] Laboratory detection of OCD can put extra faith in the mind of OCD patients. As a result, they become mentally ready for accepting OCD treatment and the treatment process becomes more effective.
	\item[Genetically Affected OCD individuals Health] As Accu-Help aims to identify GAI individuals from their OSBs. This detection is helpful to take proper precautionary measures to stay away from OCD.
	\item[Early Detection of OCD] Accu-Help also aims to predict OCD even in the early stage when symptoms are not so significant. As a result, early treatment can be recommended.
	\item[OCD Detection at Biochemical Laboratories] As Accu-Help aims to detect from OSBs, the biochemical laboratories can take help of Accu-Help to detect OCD from OSBs without involvement of a doctor.
	\item[Technological Advancement] As an OCD detector, behavioral analysis is the most commonly used method. However, this detection happens in the latter stage of the disease, and OCD patient has less faith in this detection process. As a result, they only accept treatment only at the latter stage when the disease affects their day-to-day living. The idea of using machine learning methods through an H-CPS system to analyze OSBs for the detection of HI, GAI, or OAI will be an important improvement in healthcare technology.
	
\end{description}

\section{Accu-Help: A Cyber-Physical System for Accurate Detection of OCD}
\label{sec:cps}
In recent years many cyber-physical systems are proposed to achieve smart health care \cite{rachakonda2022bactmobile}, and many more. This section proposes a conceptual cyber-physical system for organizing and managing the OCD detection process. The conceptual picture of the system is presented in Figure \ref{fig:concept}. The main component of this system is a machine learning model which can solve the OCD classification/detection problem. The efficiency of the machine learning model is mainly dependent on the labeled samples collected as training data and the learning approach. Upon completion of training and validation of the machine learning model, the model should be accessible remotely to make the model widely available and reachable to different biochemical laboratories situated in different geographical locations. Considering all these aspects in mind, the proposed health care cyber-physical system is conceptually divided into three components. The three components are 1) Labeled data collection (LDC), 2) Machine learning model design (MLMD), and 3) Machine learning model remote access (MLMRA).

\subsection{Labeled data collection (LDC)}
At the time of manual detection of OCD by a psychiatric doctor blood samples are collected from OAIs and their GAIs. The blood samples have to be sent to a biochemical laboratory for oxidative stress biomarkers (such as SD, GP, CAT, MAL, and SC) estimation. The estimated biomarkers along with their class labels have to upload to the central data cloud data server. 

\subsection{Machine learning model design (MLMD)}
Periodically, using the updated dataset machine learning model training and validation are performed in a cloud server and produce batter prediction models from time to time. The improved version of the machine learning prediction model is to be made online available for OCD detection and is called OCD prediction model. 

\subsection{Machine learning model remote access (MLMRA)}
To diagnose OCD in an individual, a blood sample of the individual is needed to be sent to a biochemical laboratory for oxidative stress biomarkers estimation. The estimated biomarkers are to be given to the OCD prediction model which is available online. The OCD prediction model in the cloud classifies the given samples into one of the three classes HI, GAI, or OAI.

\begin{figure}[tbh]
	\centering
	\includegraphics[width=0.9\textwidth]{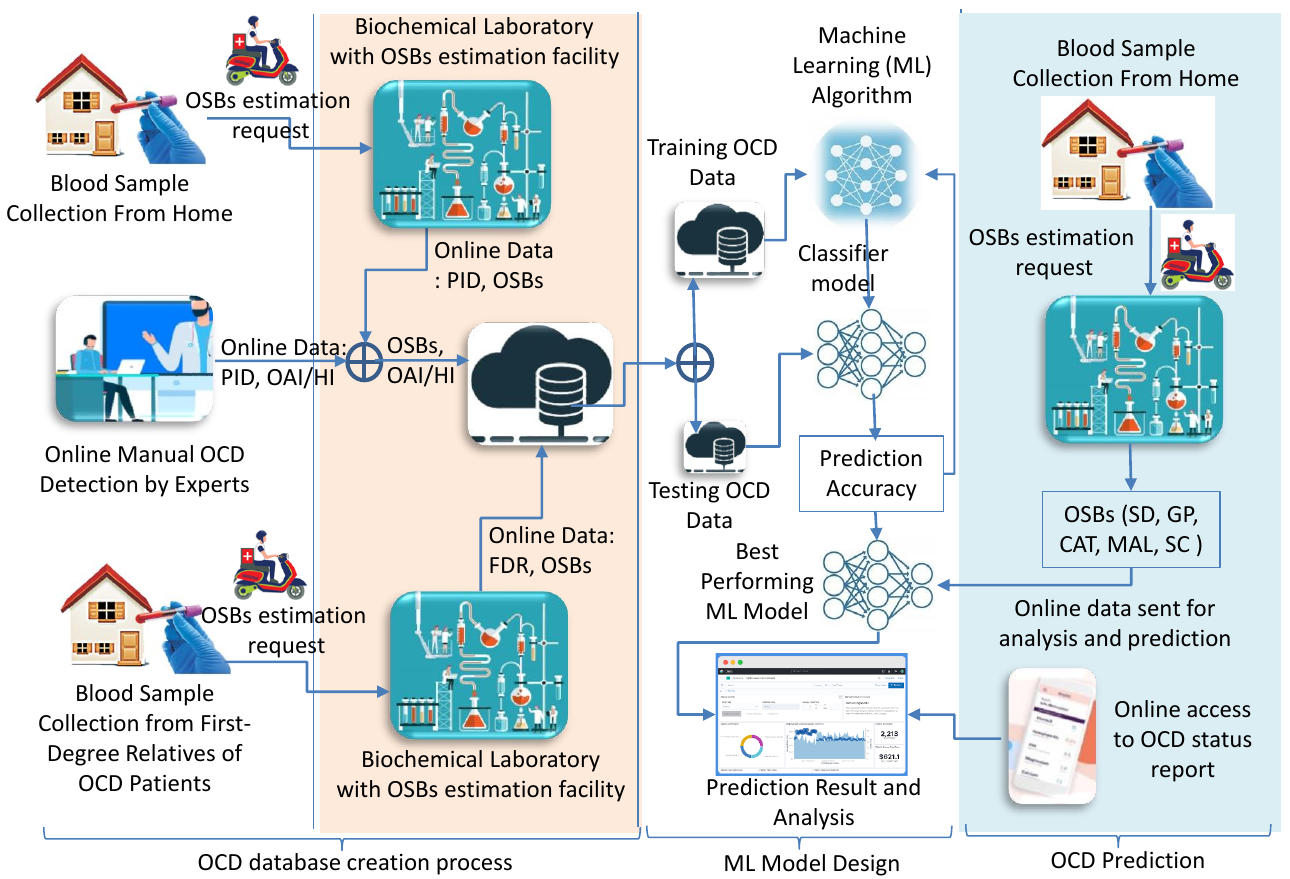}
	\caption{Conceptual Healthcare Cyber-Physical System (H-CPS) for Obsessive Compulsive Disorder (OCD) Detection}
	\label{fig:cps}
\end{figure}

The conceptual design of the cyber-physical system is presented in Figure \ref{fig:cps}. The primary objective is the construction of the OCD prediction model which is described in detail in the following sections.

\section{Classical Machine Learning Models for OCD detection}
\label{sec:classification}
\begin{figure*}[tbh]
	\centering
	\includegraphics[width=.65\textwidth]{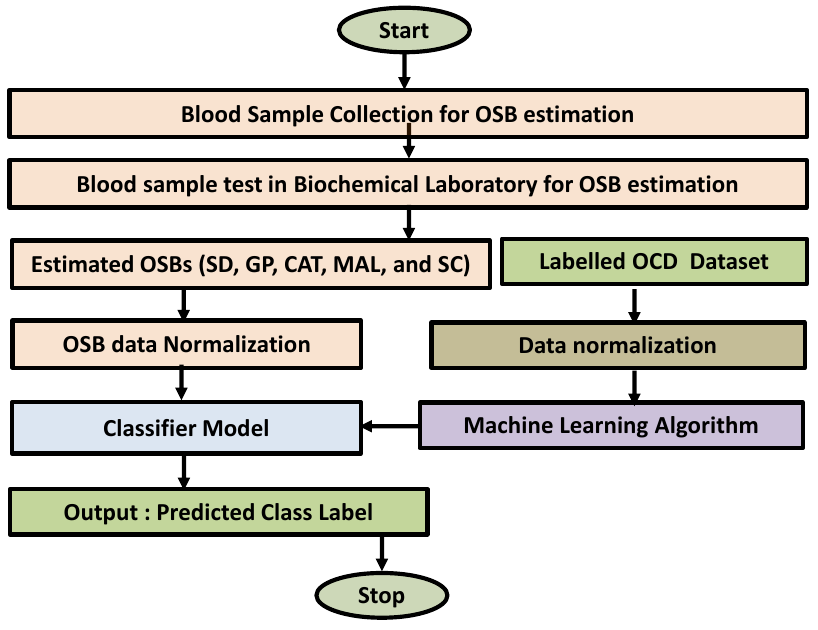}
	\caption{Proposed OCD prediction model through oxidative stress biomarkers (OSBs) analysis.}
	\label{fig:classification}
\end{figure*}

For the problem at hand, classification algorithms are suitable to solve the problem. Classification algorithms can be used to predict the class labels of unknown samples. 
The conceptual view of the OCD class prediction process is presented in Figure \ref{fig:classification}.

Among several important classifications approaches k-nearest neighbor, logistic regression, linear discriminant analysis, and neural networks are some of the popular ones. A brief description of these approaches and the OCD detection effectiveness is described in this section as follows.

\subsection{Logistic Regression and OCD Detection}
Given the OCD dataset containing $N$ number of oxidative stress biomarkers set along with the OCD class labels $\{OSBS_i, CL_i\}^N_{i=1}$. Where $CL_i\in \{HI, GAI, OAI\}$ and $OSBS_i=<SD_i, GP_i, CAT_i, MAL_i, and SC_i>$ represents the oxidative stress biomarker set of the $i^{th}$ sample containing five biomarkers called  SD, GP, CAT, MAL, and SC. The OSBs are real valued parameters  and $OSBS_i\in\Re^{5}$. The objective is to learn from the dataset and design a prediction model. For a given new $OSBS_i$ determine the class label.

\paragraph{Procedure: }Given the training dataset $\{S_{i},y_{i}\}^{N}_{i=1}$, Data point $S_{i}\in\Re^{p}$ where, $p$ is the number of predictors, class label $y_{i}\in~\{1,2,...,M)$.  The training dataset contain $N$ number of samples and the number of class levels are $M$.

\texttt{Objective:} For a given new $s\in \Re^{p}$ determine the probability of $y\in\{1,2,...,M\}$ such that $s\in class ~y$.

\texttt{Assumption:} The predictors are drawn from a probability distribution having
\begin{equation}
	Pr(y=1|s)=\frac{e^{\beta^{\tau}s}}{1+e^{\beta^{\tau}s}}=p(s;\beta)....(say),
\end{equation}
$\beta^{\tau}$ is expressed as the following:

$\beta^{\tau}=(\beta_{0},\beta_{1},\beta_{2},..., \beta_{p})$ and 
$s^{T}=(1,s_{1},s_{2},..., s_{p})$.

\begin{equation}
	\beta^{\tau}s=\beta_{0}+\beta_{1}s_{1}+\beta_{2}s_{2}+...+ \beta_{p}s_{p}
\end{equation}

For a two class classification problem,
\begin{equation}
	Pr(y=0|s)=1-p(s;\beta)
\end{equation}

$\therefore$ To know the probability that a given $s$ is from  a class $0$ or $1$ can be calculated if $\beta$ is known. $\beta$ can be calculated using the maximum likelihood method:

\begin{equation}
	Pr(y|s)=\prod_{k=1}^{N}Pr(y_{k}|s_{k})
\end{equation}

For a two class dataset:

\begin{equation}
	Pr(y|s)= \prod_{k=1}^{N}p(s_{k};\beta)^{y_{k}}(1-p(s_{k};\beta))^{1-y_{k}}
\end{equation}

where, $p(s_{k};\beta)=\frac{e^{\beta^{\tau}s_{k}}}{1+{e^{\beta^{\tau}s_{k}}}}=\frac{1}{1+{e^{-\beta^{\tau}s_{k}}}}=g(\beta^{\tau}x)$...(say)

\begin{equation}
	Let, ~~L(\beta)=Pr(y|s)=\prod_{k=1}^{N}p(s_{k};\beta)^{y_{k}}(1-p(s_{k},\beta))^{1-y_{k}}
\end{equation}

$\beta$ can be determined by maximizing the likelihood of the occurrences of all the events in the dataset.
$\therefore$ $\beta^{\tau}=(\beta_{0},\beta_{1},\beta_{2},..., \beta_{p})$ are the values for which $L(\beta)$ is maximum. But, maximizing $L(\beta)$ is same as maximizing $l(\beta)=log(L(\beta))$. To find the $\beta$ value for which $l(\beta)$ is maximum make $\triangledown l(\beta)=0$.

$\Rightarrow \frac{\partial l}{\partial \beta_{j}}=0$, for $j=0,1,...,p$

$\Rightarrow \sum_{i=1}^{N}(y_{i}-p(x_{i},\beta))x_{ij}=0$

To get $\beta$, one has to solve $\sum s_{k}(y_{k}-p(s_{k},\beta))=0$ and the solution is described in Algorithm \ref{alg:gdm}.
For a detailed description of Logistic Regression analysis one can go through \cite{field2009logistic}.

\begin{algorithm}
	\caption{$\beta$ Estimation from the given OCD dataset.}
	\label{alg:gdm}
	\textbf{Input:} OCD dataset.
	\begin{enumerate}
		\item Make initial prediction of $\beta$, lets call it $\beta^{0}$ and $k=0$
		\item $\beta^{k+1}=\beta^{k}+\alpha_{k} \triangledown l(\beta^{k})$, where $\alpha_{k}$ is a small value called learning rate.
		\item while ($\left \| \beta^{k+1}-\beta^{k} \right \|<\epsilon$)
		\item $~~~~~~\beta^{k}=\beta^{k+1}$
		\item $~~~~~~\beta^{k+1}=\beta^{k}+\alpha_{k} \triangledown l(\beta^{k})$ // Estimate $\triangledown l(\beta^{k})$ with the help of OCD dataset.
		\item Return $\beta^{k+1}$
	\end{enumerate}
\end{algorithm}

This algorithm is simulated using $R-Programming$ and experimented with OCD dataset containing OSBs and class labels. The experimental outcomes revels that this algorithm achieves  an Overall OCD classification  Accuracy of $0.777789$, Precision of  $0.768943$, Recall of $0.771247$, and F1-Score of $0.770093$.

\subsection{Linear discriminant analysis and OCD Detection}

Given the OCD dataset containing $N$ number of oxidative stress biomarkers set along with the OCD class labels $\{OSBS_i, CL_i\}^N_{i=1}$. Where $CL_i\in \{HI, GAI, OAI\}$ and $OSBS_i=<SD_i, GP_i, CAT_i, MAL_i, and SC_i>$ represents the oxidative stress biomarker set of the $i^{th}$ sample containing five biomarkers called SD, GP, CAT, MAL, and SC. The OSBs are real valued parameters  and $OSBS_i\in\Re^{5}$. The objective is to learn from the dataset and design a prediction model. For a given new $OSBS_i$ determine the class label.

\paragraph{Procedure:} Let the training dataset $D=\{S_{i},~y_{i}\}^{N}_{i=1}$, a data point $S_{i}\in\Re^{p}$ where $p$ is the number of predictors, and class label $y_{i}\in~\{1,~2,...,~M)$.  

The training dataset contains $N$ number of samples and the number of class levels is $M$.

\texttt{Objective:} For a given new sample $s\in \Re^{p}$ determine the the class of $s$.

\texttt{Classification method:} Assign the class label to x for which class probability of occurrence is maximum.
i.e.,  Class label of $s$ is $l=Arg Max_{1\leq k\leq M}~~p(y=k|S=s)$.

\begin{equation}
	p(y=k|S=s)=\frac{p(S=s|y=k)p(y=k)}{p(S=s)}
\end{equation}

This can be denoted as $p_{k}(s)=\frac{f_{k}(s)\prod_{k}}{p(s)}$, where $f_{k}(s)$ is the probability of $S=s$ given $y$, $\prod_{k}$ is the probability of $y=k$, and $p(s)$ is the probability of occurrences of $s$. It is assumed that the probability of $S=s$ for a given $y=k$ is normally distributed. i.e., $f_{k}(s)=\frac{1}{\sqrt{2\pi }\sigma_{k}}e^{\frac{-(s-\mu _{k})^{2}}{2\sigma_{k}^{2}}}$, here it is assumed that $s\in \Re$ where $\sigma_{k}$ is the standard deviation of class $k$. Again it is assumed that the standard deviation in each class are equal i.e., $\sigma_{1}= \sigma_{2}=...=\sigma_{M}=\sigma$.

\begin{equation}
	\therefore f_{k}(s)=\frac{1}{\sqrt{2\pi }\sigma}e^{\frac{-(s-\mu _{k})^{2}}{2\sigma^{2}}} 
\end{equation}

But it is known that 
$Arg Max_{1\leq k\leq M}~~p_{k}(s)=Arg Max_{1\leq k\leq M} ~~log~(p_{k}(s))$.

$\therefore$ Let us use $log~(p_{k}(s))$ to find out for which value of $k$, $p_{k}(s)$ is maximum.

$log(p_{k}(s))=log\left [ \frac{\frac{1}{\sqrt{2\pi }\sigma}e^{\frac{-(s-\mu_{k} )^{2}}{2\sigma^{2}}}\prod _{k}}{p(s)} \right ]$ 
$=log \frac{1}{\sqrt{2\pi}\sigma_{k}}-\left [ \frac{s^{2}}{2\sigma^{2}}+\frac{\mu_{k}^{2} }{2\sigma^{2}}-\frac{2s\mu_{k}}{2\sigma^{2}} \right ]+log {\prod}_{k}-log ~p(s)$.\\
$\Rightarrow Arg~Max_{1\leq k\leq M}~(p_{k}(s))=Arg~Max_{1\leq k\leq M}~(log~{\prod}_{k}-\frac{\mu_{k}^{2}}{2\sigma^{2}}+\frac{s\mu_{k}}{\sigma^{2}})$ as other terms are independent of $k$.\\
Let us denote it as 

$Arg~Max_{1\leq k\leq M}~(p_{k}(s))=Arg~Max_{1\leq k\leq M}~(\delta_{k})$

$\because$ $\prod_{k}$, $\mu_{k}$, and $\sigma$ are unknown, we can use some indirect way to estimate it. Let us define these estimated values as $\hat{\prod_{k} }=\frac{Number of elements in k^{th} claaa}{Total number of elements}=\frac{n_{k}}{N}$, $\hat{\mu_{k}}=\frac{1}{n_{k}}\sum_{i:y_{i}=k}x_{i}$ and $\hat{\sigma}=\sum_{k=1}^{M}\left ( \frac{1}{N-M}\sum_{i:y_{i}=k}(s_{i}-\hat{\mu_{k}}) \right )$.

Compute $\delta_{k}$ for all $k$ and find the $k$ for which $\delta$ is maximum. That $k$ value is the class label generated by linear discriminant analysis for $s$.

For the case where $p>1$ i.e., multiple predictors are there then the shape of linear discriminant analysis is as follows.

Suppose $s_{1}$ and $s_{2}$ are two random variables independent and normally distributed then the joint probability density function can be written as:

$f(s)=\frac{1}{\sqrt{2\pi}\sigma_{1}}e^{\frac{-(s_{1}-\mu_{1})^{2}}{2\sigma_{1}^{2}}} \frac{1}{\sqrt{2\pi}\sigma_{2}}e^{\frac{-(s_{2}-\mu_{2})^{2}}{2\sigma_{2}^{2}}}$
$=\frac{1}{2\pi\sigma_{1}\sigma_{2}}e^{-\frac{1}{2}\left [ \frac{(s_{1}-\mu_{1})^2}{\sigma_{1}^{2}} + \frac{(s_{2}-\mu_{2})^2}{\sigma_{2}^{2}} \right ]}$

In general for a $p$ random, independent and normally distributed variables/ predictors:

\begin{equation}
	f(s)=\frac{1}{2\pi^{\frac{p}{2}}\left |\sum   \right |^{\frac{p}{2}}}e^{-\frac{1}{2}(s-\mu)^\tau \sum^{-1}(x-\mu)}=p_{k}(s)
\end{equation}

where for $p=2$: $s=\begin{pmatrix} s_{1}\\ s_{2}\end{pmatrix}$, $\mu=\begin{pmatrix} \mu_{1}\\ \mu_{2}\end{pmatrix}$, and
$
{\sum}^{-1}=
\begin{pmatrix}
	\frac{1}{\sigma_1^2}& 0\\
	0 & \frac{1}{\sigma_2^2}
\end{pmatrix}
$

But we know that, 
\begin{equation}
	\begin{matrix}Arg Max\\ 1\leq k\leq M\end{matrix}~~p_{k}(s)=\begin{matrix}Arg Max\\ 1\leq k\leq M\end{matrix}~~log~(p_{k}(s))
\end{equation}

and 
\begin{equation}
	\begin{matrix}Arg Max\\ 1\leq k\leq M\end{matrix}~~log~(p_{k}(s))=\begin{matrix}Arg Max\\ 1\leq k\leq M\end{matrix}~~\delta_{k}
\end{equation}

\begin{equation}
	\label{deleq}
	where~
	\delta_k=s^\tau {\sum}^{-1}\mu_k-\frac{1}{2}\mu_k^\tau {\sum}^{-1}\mu_k+log({\prod}_k)
\end{equation}
In equation (\ref{deleq}), 
$s^\tau=(s_1, s_2, ...,s_p)$, 
$\mu_k^\tau=(\mu_1,\mu_2, ..., \mu_p)$, 

$\prod_k=\frac{Number ~of ~elements~ in~ k^{th}~ class}{Total~ number~ of~ elements}$, and
$\sum=\begin{pmatrix}
	\sigma_1^2 &0  &...   &0 \\
	0& \sigma_2^2 &...    &0 \\
	.& . &   & .\\
	.&  &  ...  & .\\
	.& . &    & .\\
	0& . &...& \sigma_p^2\end{pmatrix}$.

Classification label for $s$ is $l=\begin{matrix} Arg ~Max\\ 1\leq k\leq M \end{matrix}\delta_k$, i.e., compute $\delta_k$ for all $k$ for which $\delta$ is maximum. That $k$ value is the class label generated by linear discriminant analysis for $s$ from multi dimensional feature space. For a detailed description of Linear discriminant analysis one can go through \cite{izenman2013linear}.

This algorithm is simulated using $R-Programming$ and experimented with OCD dataset containing OSBs and class labels. The experimental outcomes revels that this algorithm achieves  an Overall OCD classification  Accuracy of $0.821431$, Precision of  $0.833333$, Recall of $0.828347$, and F1-Score of $0.830827$.

\subsection{$K$ Nearest neighbor and OCD Detection}
Given the OCD dataset containing $N$ number of oxidative stress biomarkers set along with the OCD class labels $\{OSBS_i, CL_i\}^N_{i=1}$. Where $CL_i\in \{HI, GAI, OAI\}$ and $OSBS_i=<SD_i, GP_i, CAT_i, MAL_i, and SC_i>$ represents the oxidative stress biomarker set of the $i^{th}$ sample containing five biomarkers called  SD, GP, CAT, MAL, and SC. The OSBs are real valued parameters  and $OSBS_i\in\Re^{5}$. The objective is to learn from the dataset and design a prediction model. For a given new $OSBS_i$ determine the class label.

\paragraph{Procedure:} Let the training dataset $\{S_{i},y_{i}\}^{N}_{i=1}$, Data point $S_{i}\in\Re^{p}$ where, $p$ is the number of predictors, class label $y_{i}\in~\{1,2,...,M)$.  The training dataset contain $N$ number of samples and the number of class levels are $M$.

\texttt{Objective:} For a given new $s\in \Re^{p}$ determine the the class of $s$.

The procedure of $k$-nearest neighbor is described in Algorithm \ref{alg:knn}. For a detail description of $K$ Nearest neighbor one can go through \cite{kramer2013k}.

\begin{algorithm}
	\caption{$k$ Nearest Neighbor}
	\label{alg:knn}
	\textbf{Input:} Given an OCD dataset and a new OSB sample that is to be classified.
	\begin{enumerate}
		\item Let $k$ be a positive integer and $s$ be a new sample to be classified.
		\item Evaluate  the similarity of $s$ compare to all samples of the OCD dataset using the function $distance(s, s_{j})\forall j=1, 2,..., N$. The  $distance()$ function uses euclidian distance.
		\item Sort the distances in ascending order.
		\item Consider the first $k$ distances.
		\item Identify the $k$ samples corresponding to these $k$ lowest distances.
		\item Let $c_{i}$ is the number of samples from $i^{th}$ class among these $k$ points.
		\item The new sample $s$ is classified as $c_i$ if $c_{i}>c_{j} \forall j \neq i$
	\end{enumerate}
\end{algorithm}

This algorithm is simulated using $R-Programming$ and experimented with OCD dataset containing OSBs and class labels. The experimental outcomes revels that this algorithm achieves  an Overall OCD classification  Accuracy of $0.785741$, Precision of  $0.814132$, Recall of $0.786613$, and F1-Score of $0.800102$.

\section{The Proposed Novel Hyperparameters Optimized Neural Network(HONN) for OCD Detection}
\label{sec:honn}
The neural network approach is widely adapted by researchers to solve various classification problems. However, the performance of an artificial neural network model is highly relies on the selection hyperparameters such as  number of computational unit layers in the network, activation function, learning rate, and number of computational units in each layer. In this research work, we propose an approach to optimize such hyperparameters of artificial neural network for OCD classification.

\subsection{Neural Network and OCD Detection}
Given the OCD dataset containing $N$ number of oxidative stress biomarkers set along with the OCD class labels $\{OSBS_i, CL_i\}^N_{i=1}$. Where $CL_i\in \{HI, GAI, OAI\}$ and $OSBS_i=<SD_i, GP_i, CAT_i, MAL_i, and SC_i>$ represents the oxidative stress biomarker set of the $i^{th}$ sample containing five biomarkers called  SD, GP, CAT, MAL, and SC. The OSBs are real valued parameters  and $OSBS_i\in\Re^{5}$. The objective is to learn from the dataset and design a prediction model. For a given new $OSBS_i$ determine the class label.

\begin{figure*}[htb]
	\centering
	\includegraphics[width=.8\textwidth]{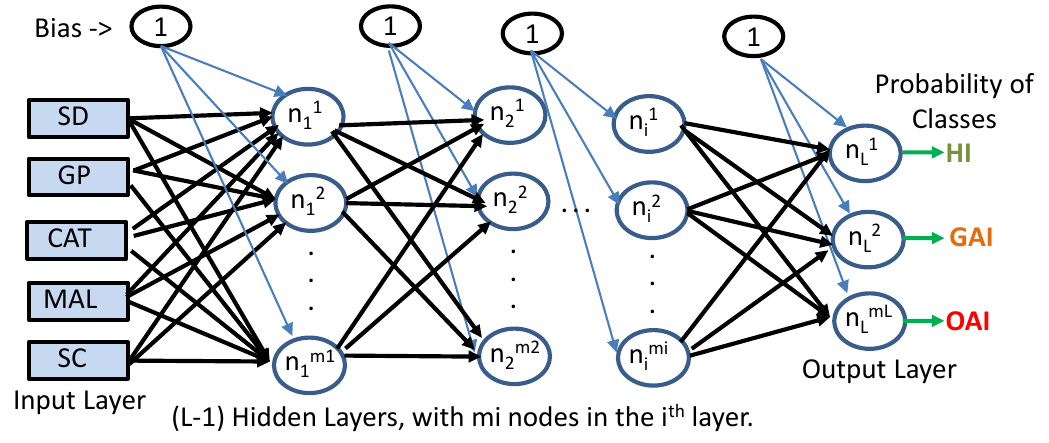}
	\caption{Proposed ANN for automatic OCD Detection}
	\label{fig:nnModel}
\end{figure*}

\paragraph{Procedure:} A generalized neural network comprises several layers of computational computational units called Input Layer (IL), zero or more Hidden Layer (HL) and Output Layer (OL).The front-end layer of computational  units is known as the IL, the backend layer of computational units is called the OL, and the computational units layers between the IL and OL  are called HL.  The IL feeds the input to the layer next to it, the computational results of the first HL computational units become the input to the next HL, and so on. Each computational unit performs a weighted sum of the inputs and passes to its activation function (AF). The AF is a continuous nonlinear function. Each OL computational units has a specific target value to produce for an associated input, the comparison of actual output and target value estimates an error signal. The error signals computed at the OL computational units and the associated weights are used to estimate the error signal at the computational units of the previous layer.
This way the error signal propagates backward layer-by-layer. The error signal is nothing but the gradient of the error function concerning the associated input weights of the computational unit. The general structure of a neural network is presented in Figure \ref{fig:nnModel}.

A set of patterns with the class labels is given as the training dataset. The training  dataset is defined by

\begin{equation}
	\label{eq1}
	TP=\{X^m, t^m\}_{m=1}^N.
\end{equation}

where $X^m\in \mathbb{R}^p$ is a  $p$ dimensional vector that represents the $m^{th}$ pattern and $t^m$ denotes the class label of $X^m$. 
Let $y_i^m$ denote the output of  computational unit $i$ at the OL as a result of the input $X^m$ at the IL.
The error signal generated at  computational unit $i$ of the OL is defined by

\begin{equation}
	\label{eq2}
	\xi_i^m=t_i^m-y_i^m
\end{equation}

where $t_i^m$ is the $i^{th}$ component of the desired response vector $t^m$.
$\therefore$ the error energy of computational unit $i$ is defined by 

\begin{equation}
	\label{eq3}
	E_i^m=\frac{1}{2}(\xi_i^m)^2
\end{equation}

Total error energy at the output layer is defined by 
\begin{equation}
	\label{eq4}
	E^m=\sum_{i} E_i^m = \frac{1}{2} \sum_{i} (\xi_i^m)
\end{equation}

The average error energy for all training samples  is defined by 
\begin{equation}
	\label{eq5}
	E_{av}=\frac{1}{N}\sum_{m=1}^{N} \sum_{i}(\xi_i^m)^2
\end{equation}

Let the computational unit $i$ being fed by  the computational units of the previous layer, The total input to computational unit $i$ is defined by 

\begin{equation}
	\label{eq6}
	v_i^m=\sum_{j=0}^{n} \omega _{ji} y_j^m
\end{equation}
where $n$ is the number of inputs (excluding the bias) to computational unit $i$. The weight $\omega_{ji}^m$ is associated with the link between $j^{th}$ computational unit of the previous layer and $i^{th}$ computational unit of the current layer. The weight $\omega_{0i}^m$  associated with the constant input $y_0=1$ represents the bias $b_i$ applied to computational unit $i$.

The activation function of the computational unit $i$ produces the output 

\begin{equation}
	\label{eq7}
	y_i^m=\psi (v_i^m)
\end{equation}

The weight correction $\Delta \omega_{ji}^m$  is applied to the weight $\omega_{ji}^m$, proportional to $\frac{\partial  E^m }{\partial \omega_{ji}^m }$.

\begin{equation}
	\label{eq8}
	\frac{\partial E^m }{\partial \omega_{ji}^m }= -\xi_i^m \psi_{i}^{'}(v_i^m)y_i^m  
\end{equation}

where $\psi_{i}^{'}(v_i^m)=\frac{\partial y_i^m}{\partial v_i^n}$, $\Delta \omega _{ji}^m=-\rho \frac{\partial E^m}{\partial \omega_{ji}^m }=\rho \gamma_i^my_j^m $, $\rho$ is the step size and $\gamma_i^m=\xi_i^m \psi_j^{'}(v_i^m)$.

For the hidden layer computational units, there is no target response. Hence, direct estimation of error signal at hidden layer computational unit is not possible.  However, the error signal of the hidden layer computational unit can be estimated using the error signal of the succeeding layer computational units and the weight associated with them. Therefore, the error signal propagates backward.

The local gradient of the error signal at a hidden layer computational unit $i$ is defined by

\begin{equation}
	\label{eq9}
	\gamma _i^m =\psi_i^{'}(v_i^m) \sum_{k} \gamma_k^m \omega_{ik}^m 
\end{equation}
where $k$ denotes a computational unit in the succeeding layer computational unit, $omega_{ik}^m$ is the weight associated between computational unit $i$ and $k$ and $\gamma_k^m$ is the local gradient of error signal at computational unit $k$. The local gradient of the error signal is computed from output layer computational units to the first hidden layer computational units in a backward direction.

$\therefore$ In general, the weight correction $\Delta \omega_{ji}^m$ formula can be defined as

\begin{equation}
	\label{eq10}
	\Delta \omega_{ji}^m=\rho \gamma_i^m y_j^m 
\end{equation}
where $\rho$ is the learning rate or step size, $\gamma_i^m$ is the local gradient and $y_j^m$ is the input signal of computational unit $j$. The computation of $\gamma_i^m$ depends on whether neuron $i$ is a hidden layer neuron or an output layer neuron.

This process of weight correction is performed several times by passing different samples each time from the training dataset until the average error comes down to an acceptable range.

The procedure for the neural network training ia described in Algorithm \ref{alg:nn}. For a detail description of the neural network training one can go through \cite{bebis1994feed}.

\begin{algorithm}
	\caption{ Network weight set learning (Training Dataset $TP$, Network Model $M$, Activation Function $\psi$, Step size $\rho$, Number of Epochs $EP$)}
	\label{alg:nn}
	\paragraph{Note:} Training Dataset $TP$ represents the oxidative stress biomarkers (SD, GP, CAT, MAL, and SC) of $N$ number of individuals.
	The network model $M(n_0, n_1,..., n_l, ..., n_L)$ represents the  $L+1$ number of layers and $n_l$ numbers of computational units in layer $l$. 
	 The input layer is represented by $l=0$ and the output layer is represented by $l=L$. 
	 The training dataset $TP=\{X^m, t^m\}_{m=1}^N$ consists of $N$ number of training patterns. The training patterns belong to $n_0$-dimensional real space, that is $X^m \in \mathbb{R}^{n_0}$. 
	 The weight $\omega_{ji}^l$ is the weight associated with the link between the $j^{th}$ computational unit of layer $l-1$ and the $i^{th}$ computational unit of layer $l$. 
	 The weight $\omega_{0i}^l$ signifies the bias of the $i^{th}$ computational unit  of the layer $l$. 
	
	\textbf{Initialization:} Initialize the weight set from a normal distribution whose mean is zero and standard deviation is a small number.

	\For{$Epoch=1$ to $EP$}{
		\For{$n=1$ to $N$}{
			Choose a random sample $X$ from $TP$ (i.e., OCD dataset) and set it as input to the network. 
			
			\For {$l=1$ to $L$}{
				\For{each computational unit $i$ in layer $l$}{ 
					$v_i^l=   \sum_{j=1}^{m_{l-1}}\omega_{ji}^l y_i^{l-1} $
					where $v_i^l$ is the input to computational unit $i$ in layer $l$, $m_{l-1}$ is the number of computational units in layer $l-1$, $y_i^{l-1}$ is the output of computational unit $i$ in layer $l-1$, and $y_i^0=X_i$.
				}
				
				$y_i^l=\psi_i^l(v_i^l)$	
			}
			\For{$i=1$ to $m_L$}{
				$\xi_i^n=t_i-y_i^L$
			}
			\For{$i=1$ to $m_L$}{
				$\gamma_i^L=\xi_i {\psi_i^l}'(v_i^L)$
			}
			\For {$l=L-1$ to $1$ }{
				\For{$i=1$ to $m_l$}{
					${\psi_i^l}'(v_i^l)\sum_{k=1}^{m_{l+1}}\gamma_k^{l+1}\omega_{ik}^{l+1}$
				}
			}
			\For {$l=1$ to $L$ }{
				\For{$j=0$ to $m_{l-1}$}{
					\For{$i=1$ to $m_l$}{
						$\Delta \omega_{ji}=\rho \gamma_i^l y_j^{l-1} $
						
						$\omega_{ji}=\omega_{ji}+\Delta \omega_{ji}$
					}
				}
			}
		}
		\For{$n=1$ to $N$}{
			Using $X^n$ as input calculate $y_n^L$ and estimate $\xi_n$.}

		$\xi_{av}=\frac{1}{N}\sum_{n=1}^{N}\xi_n$
		
	}
	
	\textbf{Return}(Weight set $W$)
	
\end{algorithm}

This algorithm is simulated using $R-Programming$ and experimented with OCD dataset containing OSBs and class labels. The experimental outcomes revels that this algorithm achieves  an Overall OCD classification  Accuracy of $0.833333$, Precision of  $0.839907$, Recall of $0.83482$, and F1-Score of $0.836035$.

\subsection{Hyperparameter Optimization Procedure}
The OCD class prediction accuracy by a neural network mainly depends on various hyperparameters.   
It is crucial to find optimal hyperparameters to increase the performance of a network. The approach proposed in this article adopts a finite hyperparameter set guestimating approach called HONN.

\begin{figure}[htb]
	\centering
	\includegraphics[width=.7\textwidth]{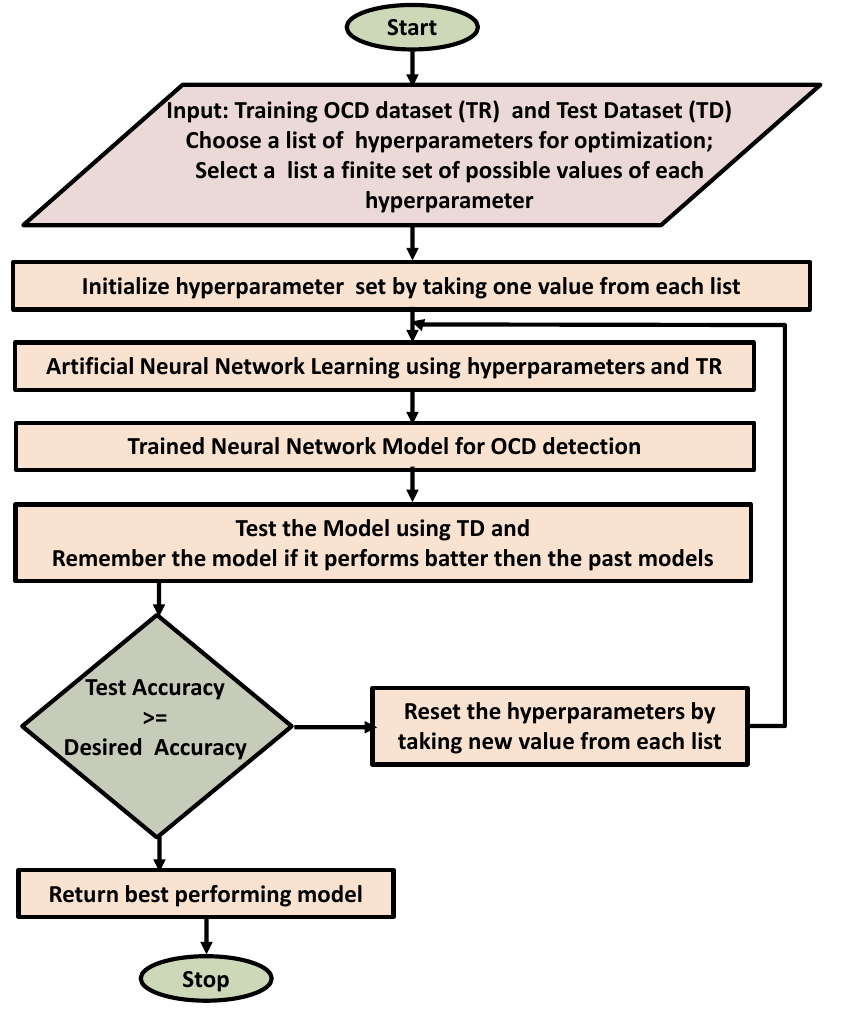}
	\caption{Proposed flow for novel Hyperparameters Optimized Neural Network (HONN) modeling.}
	\label{fig:honn}
\end{figure}

\paragraph{HONN ModelArchitecture}
Select a set of hyperparameters that needs to be optimized. For each hyperparameter choose a finite list of possible values. Initialize the hyperparameters of the neural network model by taking one value from the lists of each parameters list. Perform the neural network training by providing the training dataset. Once the neural network is trained, test it with the help of a test dataset and record the test accuracy. If the accuracy achieved in this step is better than the past models then preserve the parameters set. Repeat this process of new hyperparameter set selection from the list, training, and testing process for all combinations of hyperparameter sets. Finally, return the parameters that are preserved as best-performing ones. The conceptual architecture of the HONN is presented in Figure \ref{fig:honn}.    

Out of many hyperparameters, this approach tries to optimize the activation function, the number of layers in the network, the number of computational units in each layer, and number of epoch. 
The activation functions considered for different computational units of the neural network are Logistic $\left(f(v)=\frac{1}{1+e^{-v}} \right)$,  Tanh    $\left (f(v)=\frac{2}{1+e^{-2v}}-1 \right )$, ArcTan $\left(tan^{-1}(v) \right)$, and Softplus $\left (log_e(1+e^{v}) \right)$. Apart from the input and output layer number of layers considered  are $0$, $1$, and $2$. The number of computational units in each layer ranges from $3$ to $15$. The hyperparameter optimization or model selection approach is presented in Algorithm \ref{alg:honn}.

\begin{algorithm}
	\caption{ Hyperparameter optimization neural network algorithm (Training Dataset $TP$, Test Dataset $TD$)}
	\label{alg:honn}
	\paragraph{Input:}The OCD dataset is divided into training and test dataset (i.e., $TP$ and $TD$). $TP$ is utilized to train the network where as $TD$ is used to estimate the accuracy of the trained model.\\
	\textbf{Initialization:} \\	
	AF=[Logistic, Tanh, ArcTan, Softplus] //Activation Function list\\
	SS=[$\rho_1$, $\rho_2$, ..., $\rho_m$] // Step size list\\
	EL=[$EP_1$,$EP_1$, ..., $EP_n$]	//Number of Epoch list\\
	$Accuracy=0$, $Model=(5,3)$\\
	\textbf{Procedure:}\\
	\For {$f$ in AF}{
		\For{$\rho$ in SS}{
			\For{$EP$ in EL}{
				\For{$l=0$ to $2$}{
					\If{$l==0$}{
						$M=(5,3)$\\
						$W$=Algorithm \ref{alg:nn}($TP,~M, ~f, ~\rho, EP$ )\\
						$newAccuracy=ModelAccuracyTest(M, $W$, TD)$\\
						\If{$newAccuracy>Accuracy$}{
							$Accuracy=newAccuracy$, $Model=M$, Preserve $W$.
						}
					}
					\eIf{$l==1$}{
						\For{$i=3$ to $15$}{
							$M=(5,i,3)$\\
							$W$=Algorithm \ref{alg:nn}($TP,~M, ~f, ~\rho, EP$ )
							$newAccuracy=ModelAccuracyTest(M, $W$, TD)$\\
							\If{$newAccuracy>Accuracy$}{
								$Accuracy=newAccuracy$, $Model=M$, Preserve $W$.
							}
						}
					}
					{
						\For {$j=3$ to $15$}{
							\For{$i=3$ to $15$}{
								$M=(5,i,j,3)$
								$W$=Algorithm \ref{alg:nn}($TP,~M, ~f, ~\rho, EP$ )
								$newAccuracy=ModelAccuracyTest(M, $W$, TD)$\\
								\If{$newAccuracy>Accuracy$}{
									$Accuracy=newAccuracy$, $Model=M$, Preserve $W$.
								}
							}
						}
						
					}
					
				}
			}
		}
	}
	\textbf{Return}(Model $M$, and Weight Set $W$)
\end{algorithm}

This article uses KNN, LR, LDA, ANN and HONN for OSBs classification for OCD class detection. The working principle comparative study of these approach are presented in Table \ref{tab:approachCompe}.

\begin{table}[tbh]
	\centering
	\caption{OCD classification approaches feature study.}
	\label{tab:approachCompe}
	\begin{tabular}{|l|p{0.7\textwidth}	|}
		\hline
		\textbf{Approach} & \textbf{Working Principles and reasons for the performance}                                                                                     \\ 
		& \\ \hline
		KNN               & This approach classifies the new samples based on the local majority class of the new OSB sample.                                                                             \\ \hline
		LR                & The class label prediction is performed based on the probability of the sample belonging to a particular class. The higher probable class is selected.                        \\ \hline
		LDA               & This approach works well if the features follow normal distribution.                                                                                                          \\ \hline
		ANN               & ANN is quite suitable to handle linearly nonseparable classes. However, the technique is not performing its best if the approach's hyperparameters are not suitably selected. \\ \hline
		HONN              & ANN HONN tries to improve the ANN approach by proposing a parameter selection algorithm.                                                                                      \\ \hline
	\end{tabular}
\end{table}

\section{Experimental Evaluation}
\label{sec:exp}
A quantitative analysis of the effectiveness of the proposed approach is performed by experimenting with the proposed approach and testing with the OCD data which was originally presented in \cite{kar2016empirical}. To make a comparative analysis the performance of the approach is compared with the performance of other popular and relevant approaches by considering the same dataset as input. Among the state-of-the-art supervised approaches, some of the most popular supervised approaches such as \emph{k-nearest neighbor, logistic regression, linear discriminant analysis, and neural network} are considered. A brief description of the dataset is given in the following paragraph. 

\subsection{Oxidative Stress Biomarker Dataset Description}
%\label{sec:data}
Oxidative stress biomarkers are recorded from the blood samples of healthy individuals, OCD patients, and first-degree relatives of OCD patients. The restriction imposed during individual selection for sample collection includes: age should be between $18$ and $45$, should not be under any medications in the last three months, should not have any illness in case of healthy and first-degree relatives, and should not have any other illness in case of OCD patients. Pregnant and lactating individuals are excluded. The oxidative stress biomarkers considered for this study are  SD, GP, CAT, MAL, and SC. Standard biochemical methods are adopted to measure these biomarkers. To estimate these biomarkers, blood samples are drawn after overnight fasting. The assessment of these biomarkers is carried out in the biochemical laboratory of King George Medical University. Plasma is used as the source of enzymes. An in-depth description of the laboratory mechanism of assessment of these biomarkers is presented in \cite{kar2016empirical}.

\begin{figure*}[hbt!]
	\centering
	\includegraphics[width=\textwidth]{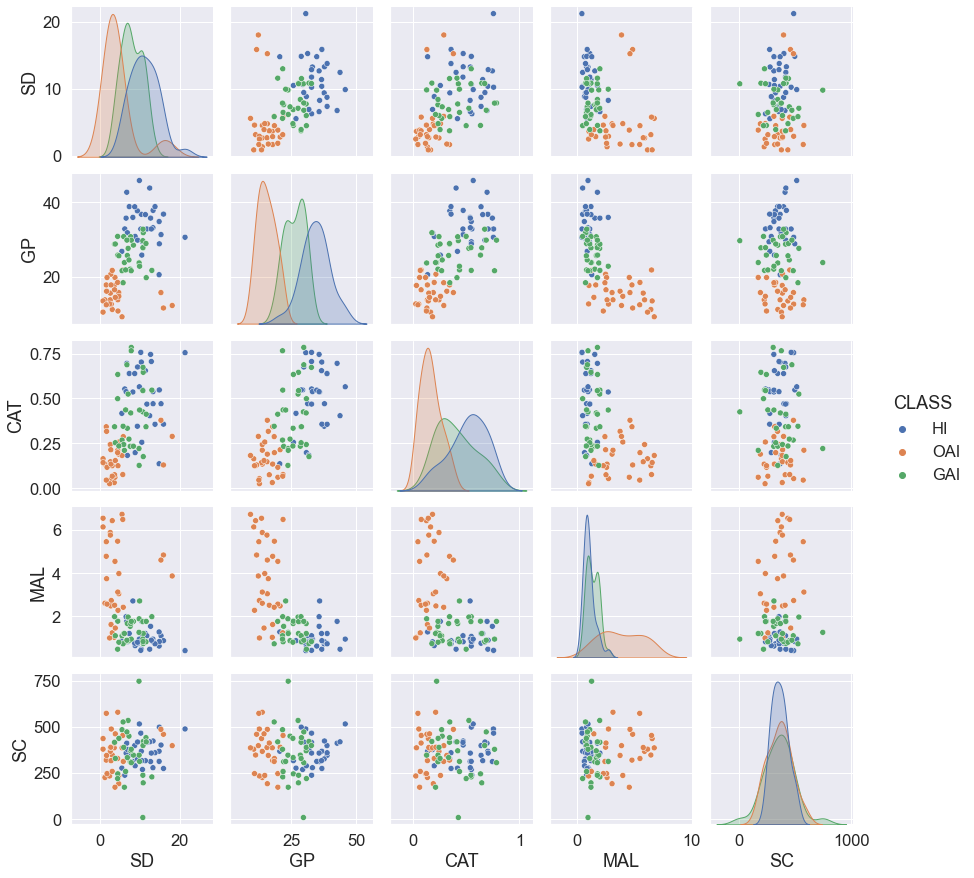}
	\caption{Class density distribution over biomarkers and all biomarker pair scatter plots }
	\label{fig:scatterPlot}
\end{figure*}

The biomarkers recorded are in the form of real numbers. The range of the values varies from marker to marker. The distribution of the values in different metrics  and all biomarker pair wise scatter plot is presented in Figure \ref{fig:scatterPlot}.

To project all the values of the markers into a common range a normalization procedure is followed. 
Let $\beta=\{\beta_1, \beta_2, ..., \beta_p\}$ represents the set of biomarkers where $p$ is the number of biomarkers.

Let $\beta_i=\{\beta_i^1, \beta_i^2,..., \beta_i^N\}$ be the collection of $i^{th}$ marker for all $N$ samples.
Let $\beta_i^s$ and $\beta_i^l$ represent  the smallest and largest value among the values stored in $\beta_i$ respectively.
To normalize $\beta_i$ in the range $[n_1, n_2]$, we adopt the min-max normalization method. The normalized value $\overline{\beta_i^j}=n_1+(n_2-n_1)\left ( \frac{\beta_i^j-\beta_i^s}{\beta_i^l-\beta_i^s} \right )$.

\subsection{Quantitative Analysis}
To access the effectiveness of the model, a test dataset $TD=\{X_i,y_i\}_{i=1}^M$ is used which consists of $M$ number of samples and these samples are not used for the training of the model. 

$X_i=<X_i^{SD},~ X_i^{GP},~ X_i^{CAT},~X_i^{MAL},~X_i^{SC} >$ represents the five oxidative stress biomarkers of the  $i^{th}$ sample of the test dataset and $y_i$ is the class label of $X_i$. The test samples ($X_i$) are presented to the model to get the predicted output $\hat{y_i}$ of a model and $\hat{y_i}$ is compared with $y_i$ to get the test  accuracy of the model. The probable outcomes of this comparison is given in equation \ref{eq11} where $TP_c$ represents the true positive for class $c$, $TN_c$ represents the true negative for class $c$, $FP_c$ represents the false positive for class $c$, and $FN_c$ represents the true positive for class $c$,

\begin{equation}
	\label{eq11}
	Comp(\hat{y_i},y_i)\in \\
	\left\{
	\begin{matrix}
		TP_c &~if~ \hat{y_i}=y_i= c ~(class~ label~)\\ 
		TN_c &~if~ \hat{y_i}~and~y_i \neq   c\\
		FP_c &~if~ \hat{y_i}=c~and~y_i \neq   c\\
		FN_c &~if~ \hat{y_i}\neq c~ and~ y_i= c\\
	\end{matrix}
	\right.
\end{equation}

The output of the $Comparison$ function for all the samples can be represented in a confusion matrix. The confusion matrix $CM$ for a $3$ class classification problem can be represented as a $(3 \times 3)$ matrix, where $CM[i,j]$ represents the number of samples predicted as class $i$ and the actual class of the sample is $j$. $\therefore$ $TP_1=CM[1,1]$, $TP_2=CM[2,2]$, $TP_3=CM[3,3]$,
$TN_1=CM[2,2]+CM[2,3]+CM[3,2]+CM[3,3]$, 
$TN_2=CM[1,1]+CM[1,3]+CM[3,1]+CM[3,3]$, 
$TN_3=CM[1,1]+CM[1,2]+CM[2,1]+CM[2,2]$, 
$PF_1=CM[1,2]+CM[1,3]$,
$PF_2=CM[2,1]+CM[2,3]$,
$PF_3=CM[3,1]+CM[3,2]$,
$FN_1=CM[2,1]+CM[3,1]$,
$FN_2=CM[1,2]+CM[3,2]$, and
$FN_3=CM[1,3]+CM[2,3]$.

Based on the values of $TP,~ TN,~ FP~$, and $FN$ of all the classes, one can estimate classification accuracy metrics such as \emph{Overall Accuracy,  Precision, Recall, and  F1-Score}. The formulas used to estimate these metrics are given in Table \ref{tab:metrics}.

\begin{table}[]
	\centering
	\caption{Classification accuracy metrics.}
	\label{tab:metrics}

	\begin{tabular}{|l|l|}
		\hline
		\textbf{Accuracy Metric}		 	& \textbf{Estimation Method} \\ 
		 & \\ \hline
		\emph{Overall Accuracy}			 	& $\left(\sum_{c=1}^{C}\frac{TP_c+TN_c}{TP_c+FN_c+FP_c+TN_c}\right)/3$ \\ \hline
		\emph{Precision}			 		& $\frac{\sum_{c=1}^{C}TP_c}{\sum_{c=1}^{C}(TP_c+FP_c)}$ \\ \hline
		\emph{Recall}				  		& $\frac{\sum_{c=1}^{C}TP_c}{\sum_{c=1}^{C}(TP_c+FN_c)}$ \\ \hline
		\emph{F1-Score}				 		& $2*\left(\frac{Precision*Recall}{Precision+Recall}\right)$ \\ \hline
	\end{tabular}
\end{table} 

\subsection{Experimental Setup}
For a better conformation of the model accuracy a $k-fold$ cross-validation approach is adopted. $k$ round, of experiments are performed and in each round the model accuracy is estimated. The average over all the $k$ experiments accuracy results is considered as the model accuracy. In the $k-fold$ cross-validation approach, the given OCD dataset is randomly shuffled and divided into $k$ nearly equal size partitions. That is the given dataset $D=\{D_1, D_2,...,D_k\}$. In the $i^{th}$ round of the experiment, the training dataset $TP=D_1 \cup D_2 \cup...\cup D_{i-1} \cup D_{i+1} \cup D_k$ and $TD=D_i$. Due to the small size of the dataset, in the current study, a $3-fold$ cross-validation process is repeated multiple times with a complete data shuffle and re-partition after each $3-fold$ cross-validation to achieve  $10-fold$ cross-validation. The experiments are performed on a system having \texttt{Intel(R) Core(TM) i5-3210m, 4GB RAM, 2.0GHz Processor and Windows-11 as OS}. \texttt{R-4.2.1} is used for programming the approaches.

\subsection{Comparative Analysis}
For comparative analysis, the experiments are performed twofold. In the first fold, manually different variants of neural networks are selected and experimented with for OCD classification and the outcomes are compared with the outcomes of the proposed model.
In the second fold of the experiment, some of the popular classification approaches (such as k-Nearest Neighbor, Logistic Regression, Linear Discriminant Analysis, and Neural Network) are selected and experimented with for OCD classification and the outcomes are compared with the outcomes of the proposed model.

\begin{table}
	\centering
	\caption {Classification accuracy achieved by the neural network models: Mean values of all accuracy measures over 10-fold validation (HLN: Hidden Layer Numbers, and HLNN: Hidden Layer computational unit Numbers)}
	\label{tab:nnAccuracy}

	\begin{tabular}{|l|l|l|l|}
		\hline
		\multicolumn{1}{|l|}{\textbf{Accuracy measure}} &\multicolumn{1}{l|}{ \textbf{HLN}}       &\multicolumn{1}{l|}{ \textbf{HLNN} }        &\multicolumn{1}{l|}{ \textbf{Accuracy}} \\
		& & & \\		\hline
		\multicolumn{1}{|l|}{Overall Accuracy} &\multicolumn{1}{l|}{ \multirow{4}{*}{1}} & \multicolumn{1}{l|}{\multirow{4}{*}{6} }      & \multicolumn{1}{l|}{0.833333} \\\cline{1-1} \cline{4-4}
		\multicolumn{1}{|l|}{Precision }      & \multicolumn{1}{l|}{}                   & \multicolumn{1}{l|}{}                          & \multicolumn{1}{l|}{0.839907} \\\cline{1-1} \cline{4-4}
		\multicolumn{1}{|l|}{Recall}           & \multicolumn{1}{l|}{}                  &  \multicolumn{1}{l|}{}                         & \multicolumn{1}{l|}{0.83482}  \\\cline{1-1} \cline{4-4}
		\multicolumn{1}{|l|}{F1 score }       &  \multicolumn{1}{l|} {}               &  \multicolumn{1}{l|}{}                         &\multicolumn{1}{l|}{0.836035} \\\hline
		\multicolumn{1}{|l|}{Overall Accuracy} &\multicolumn{1}{l|}{ \multirow{4}{*}{1}} & \multicolumn{1}{l|}{\multirow{4}{*}{10} }      & \multicolumn{1}{l|}{0.777778} \\\cline{1-1} \cline{4-4}
		\multicolumn{1}{|l|}{Precision }      & \multicolumn{1}{l|}{}                   & \multicolumn{1}{l|}{}                          & \multicolumn{1}{l|}{0.768943} \\\cline{1-1} \cline{4-4}
		\multicolumn{1}{|l|}{Recall}           & \multicolumn{1}{l|}{}                  &  \multicolumn{1}{l|}{}                         & \multicolumn{1}{l|}{0.771247}  \\\cline{1-1} \cline{4-4}
		\multicolumn{1}{|l|}{F1 score }       &  \multicolumn{1}{l|} {}               &  \multicolumn{1}{l|}{}                         &\multicolumn{1}{l|}{0.770093} \\\hline
		\multicolumn{1}{|l|}{Overall Accuracy} &\multicolumn{1}{l|}{ \multirow{4}{*}{1}} & \multicolumn{1}{l|}{\multirow{4}{*}{15} }      & \multicolumn{1}{l|}{0.811111} \\\cline{1-1} \cline{4-4}
		\multicolumn{1}{|l|}{Precision }      & \multicolumn{1}{l|}{}                   & \multicolumn{1}{l|}{}                          & \multicolumn{1}{l|}{0.825838} \\\cline{1-1} \cline{4-4}
		\multicolumn{1}{|l|}{Recall}           & \multicolumn{1}{l|}{}                  &  \multicolumn{1}{l|}{}                         & \multicolumn{1}{l|}{0.820978}  \\\cline{1-1} \cline{4-4}
		\multicolumn{1}{|l|}{F1 score }       &  \multicolumn{1}{l|} {}               &  \multicolumn{1}{l|}{}                         &\multicolumn{1}{l|}{0.823345} \\\hline
		\multicolumn{1}{|l|}{Overall Accuracy} &\multicolumn{1}{l|}{ \multirow{4}{*}{2}} & \multicolumn{1}{l|}{\multirow{4}{*}{5,5} }      & \multicolumn{1}{l|}{0.755556} \\\cline{1-1} \cline{4-4}
		\multicolumn{1}{|l|}{Precision }      & \multicolumn{1}{l|}{}                   & \multicolumn{1}{l|}{}                          & \multicolumn{1}{l|}{0.769639} \\\cline{1-1} \cline{4-4}
		\multicolumn{1}{|l|}{Recall}           & \multicolumn{1}{l|}{}                  &  \multicolumn{1}{l|}{}                         & \multicolumn{1}{l|}{0.747397}  \\\cline{1-1} \cline{4-4}
		\multicolumn{1}{|l|}{F1 score }       &  \multicolumn{1}{l|} {}               &  \multicolumn{1}{l|}{}                         &\multicolumn{1}{l|}{0.75834} \\\hline
		\multicolumn{1}{|l|}{Overall Accuracy} &\multicolumn{1}{l|}{ \multirow{4}{*}{2}} & \multicolumn{1}{l|}{\multirow{4}{*}{10,8} }      & \multicolumn{1}{l|}{0.788889} \\\cline{1-1} \cline{4-4}
		\multicolumn{1}{|l|}{Precision }      & \multicolumn{1}{l|}{}                   & \multicolumn{1}{l|}{}                          & \multicolumn{1}{l|}{0.800385} \\\cline{1-1} \cline{4-4}
		\multicolumn{1}{|l|}{Recall}           & \multicolumn{1}{l|}{}                  &  \multicolumn{1}{l|}{}                         & \multicolumn{1}{l|}{0.773107}  \\\cline{1-1} \cline{4-4}
		\multicolumn{1}{|l|}{F1 score }       &  \multicolumn{1}{l|} {}               &  \multicolumn{1}{l|}{}                         &\multicolumn{1}{l|}{0.786435} \\\hline
		\multicolumn{1}{|l|}{Overall Accuracy} &\multicolumn{1}{l|}{ \multirow{4}{*}{2}} & \multicolumn{1}{l|}{\multirow{4}{*}{15,10} }      & \multicolumn{1}{l|}{0.733333} \\\cline{1-1} \cline{4-4}
		\multicolumn{1}{|l|}{Precision }      & \multicolumn{1}{l|}{}                   & \multicolumn{1}{l|}{}                          & \multicolumn{1}{l|}{0.74236} \\\cline{1-1} \cline{4-4}
		\multicolumn{1}{|l|}{Recall}           & \multicolumn{1}{l|}{}                  &  \multicolumn{1}{l|}{}                         & \multicolumn{1}{l|}{0.738571}  \\\cline{1-1} \cline{4-4}
		\multicolumn{1}{|l|}{F1 score }       &  \multicolumn{1}{l|} {}               &  \multicolumn{1}{l|}{}                         &\multicolumn{1}{l|}{0.740436} \\\hline
		\multicolumn{1}{|l|}{Overall Accuracy} &\multicolumn{1}{l|}{ \multirow{4}{*}{3}} & \multicolumn{1}{l|}{\multirow{4}{*}{5,5,4} }      & \multicolumn{1}{l|}{0.7} \\\cline{1-1} \cline{4-4}
		\multicolumn{1}{|l|}{Precision }      & \multicolumn{1}{l|}{}                   & \multicolumn{1}{l|}{}                          & \multicolumn{1}{l|}{0.705804} \\\cline{1-1} \cline{4-4}
		\multicolumn{1}{|l|}{Recall}           & \multicolumn{1}{l|}{}                  &  \multicolumn{1}{l|}{}                         & \multicolumn{1}{l|}{0.686802}  \\\cline{1-1} \cline{4-4}
		\multicolumn{1}{|l|}{F1 score }       &  \multicolumn{1}{l|} {}               &  \multicolumn{1}{l|}{}                         &\multicolumn{1}{l|}{0.695818} \\\hline
		\multicolumn{1}{|l|}{Overall Accuracy} &\multicolumn{1}{l|}{ \multirow{4}{*}{3}} & \multicolumn{1}{l|}{\multirow{4}{*}{10,8,5} }      & \multicolumn{1}{l|}{0.655556} \\\cline{1-1} \cline{4-4}
		\multicolumn{1}{|l|}{Precision }      & \multicolumn{1}{l|}{}                   & \multicolumn{1}{l|}{}                          & \multicolumn{1}{l|}{0.668515} \\\cline{1-1} \cline{4-4}
		\multicolumn{1}{|l|}{Recall}           & \multicolumn{1}{l|}{}                  &  \multicolumn{1}{l|}{}                         & \multicolumn{1}{l|}{0.665826}  \\\cline{1-1} \cline{4-4}
		\multicolumn{1}{|l|}{F1 score }       &  \multicolumn{1}{l|} {}               &  \multicolumn{1}{l|}{}                         &\multicolumn{1}{l|}{0.666983} \\\hline
		\multicolumn{1}{|l|}{Overall Accuracy} &\multicolumn{1}{l|}{ \multirow{4}{*}{3}} & \multicolumn{1}{l|}{\multirow{4}{*}{15,10,8} }      & \multicolumn{1}{l|}{0.722222} \\\cline{1-1} \cline{4-4}
		\multicolumn{1}{|l|}{Precision }      & \multicolumn{1}{l|}{}                   & \multicolumn{1}{l|}{}                          & \multicolumn{1}{l|}{0.738492} \\\cline{1-1} \cline{4-4}
		\multicolumn{1}{|l|}{Recall}           & \multicolumn{1}{l|}{}                  &  \multicolumn{1}{l|}{}                         & \multicolumn{1}{l|}{0.731519}  \\\cline{1-1} \cline{4-4}
		\multicolumn{1}{|l|}{F1 score }       &  \multicolumn{1}{l|} {}               &  \multicolumn{1}{l|}{}                         &\multicolumn{1}{l|}{0.734933} \\\hline
	\end{tabular}
\end{table}

\begin{figure*}[tbh!]
	\centering
	\includegraphics[width=\textwidth]{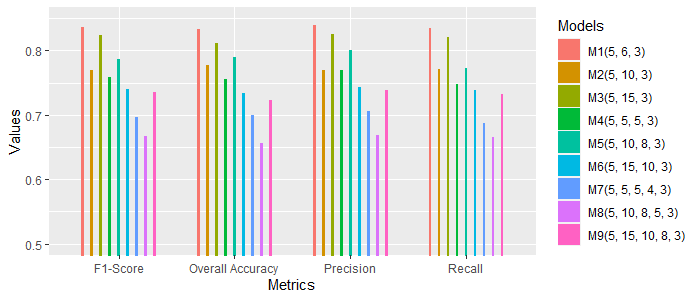}
	\caption{Classification accuracy comparison bar plots for all network models}
	\label{fig:nnBarPlots}
\end{figure*}

\begin{table*}[]
	\centering
	\caption{Classification accuracy achieved by ANN, KNN, LR, LDA, and HONN : Mean values of all accuracy measures over 10-fold validation}
	\label{tab:approachComparison}

	\begin{tabular}{|l|l|l|l|l|l|}
		\hline
		\textbf{Metric}           & \textbf{ANN} & \textbf{KNN} & \textbf{LR} & \textbf{LDA} & \textbf{HONN} \\ 
		& & & & & \\ \hline
		\textbf{Overall Accuracy} & 0.833333                & 0.785741                    & 0.777789                     & 0.821431                              & 0.861111      \\ \hline
		\textbf{Precision}        & 0.839907                & 0.814132                    & 0.768943                     & 0.833333                              & 0.8650794     \\ \hline
		\textbf{Recall}           & 0.83482                 & 0.786613                    & 0.771247                     & 0.828347                              & 0.8630752     \\ \hline
		\textbf{F1-Score}         & 0.836035                & 0.800102                    & 0.770093                     & 0.830827                              & 0.8640761     \\ \hline
	\end{tabular}
\end{table*}

\begin{figure*}[hbt!]
	\centering
	\includegraphics[width=12cm]{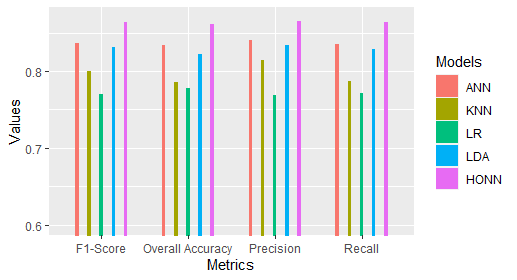}
	\caption{Classification accuracy comparison bar plots for ANN, KNN, LR, LDA, and HONN:: Mean values of all accuracy measures over 10-fold validation}
	\label{fig:allBarPlots}
\end{figure*}

Nine different neural network architectures are selected for OCD classification. The models which are chosen are defined by 
$M1(5,~6,~ 3)$, 
$M2(5,~10,~3)$, 
$M3(5,~15,~3)$, 
$M4(5,~5,~5,3)$, 
$M5(5,~10,~8,~ 3)$, 
$M6(5,~ 15,~ 10,~ 3)$,
$M7(5,~5,~5,~4,~3)$, 
$M8(5,~10,~8,~5,~3)$, and
$M9(5,~15,~10,~8,~3)$.  
Logistic function is used as the activation function for all the computational units. The step size or learning rate is set to $0.005$ and the maximum number of epoch is taken as $10000$. All these models goes through a $10-fold$ cross-validation process. The \emph{Overall Accuracy, Error Rate, Precision, Recall, Micro Averaging F1-Score, Macro Averaging F1-Score} obtained by different models are presented in Table \ref{tab:nnAccuracy}.
For visual analysis, bar plots are used for each accuracy measure. The \emph{x-axis} of these plots is marked with the metrics, the \emph{y-axis}
is marked with the accuracy level, and different colours are used to represent different models. The bar plots are presented in Figure \ref{fig:nnBarPlots}. From Table \ref{tab:nnAccuracy} and Figure \ref{fig:nnBarPlots} it can be observed that the model $M1$ performs batter compared to others with an \emph{Overall Accuracy} of $0.833333$, \emph{Precision} of $0.839907$, \emph{Recall} of $0.83482$ and \emph{F1-Score} of $0.836035$.

Further, experiments are performed using \emph{k-Nearest Neighbor (KNN), Logistic Regression (LR), Linear Discriminant Analysis (LDA), Neural Network (Model M1) (ANN), and HONN} for OCD classification. The hyperparameters of these models are tuned using hit-and-trial method. $10-fold$ cross-validation process is followed to estimate performance accuracy. For a comparative analysis, the results  obtained are presented in Table \ref{tab:approachComparison}. Accuracy measure wise bar plots are presented in Figure \ref{fig:allBarPlots} for a visual comparative analysis. From Table \ref{tab:approachComparison} and Figure \ref{fig:allBarPlots} it can be observed that \emph{HONN} performs batter compared to others with an \emph{Overall Accuracy} of $0.861111$, \emph{Precision} of $0.8650794$, \emph{Recall} of $0.8630752$ and \emph{F1-Score} of $0.8640761$.

\section{Conclusion and Future Directions of Research}
\label{sec:con}
In this study, a hyperparameter-optimized neural network (HONN) approach is proposed to classify oxidative stress biomarkers into one   category  from HI, GAI, and OAI. Instead  making a hit-and-trial method to achieve an optimal neural network model is quite hard. The classification accuracy depends on the hyperparameters set for the network. This challenge can be reduced by adopting the HONN method. For comparative study \emph{k-Nearest Neighbor, Logistic Regression, Linear Discriminant Analysis, and Artificial Neural Network } are used. From experimental result  analysis, it is observed that HONN yields better accuracy in the test dataset with respect to all the classification measures. The classification accuracy obtained by HONN is $86\pm 2\%$. The identification of GAI indicates the genetic  linkage to OCD. In such cases, appropriate preventive actions can be recommended well in advance. Data collection, class label integration, machine learning model design and online OCD classification process can be managed and monitored using Accu-Help.

Along with expanding the scope of the Accu-Help and OCD detection process more accurate can be our future research directions. Even though the proposed machine learning model performs well compared to simple models, the approach is not scalable. This approach is computationally costly. In the future, we will try to propose a scalable model with better OCD detection accuracy.

\section*{Acknowledgment}
This work is supported by the Odisha Higher Education Programme for Excellence and Equity (OHEPEE) World Bank [No. 6770/GMU].
%%%%%%%%%%%%%%%%%%%%%%%%%%%%%%%%%%%%%%%%%%%%%%%%%%%%%

\bibliographystyle{IEEEtran}	
\bibliography{myBibDB}

\begin{wrapfigure}[10]{l}{35 mm} 
	\includegraphics[height=1.5in, width=1.25in]{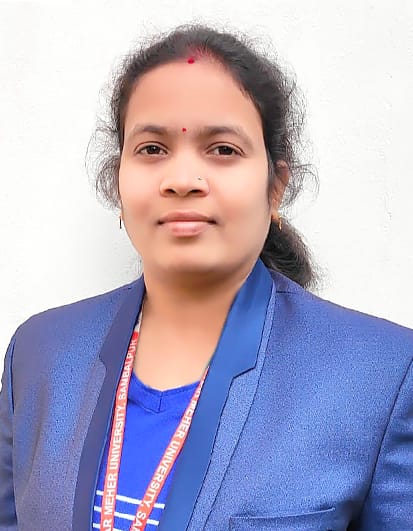}
\end{wrapfigure}\par
\textbf{Kabita Patel}  completed the B. Sc. degree in computer science from Gangadhar Meher University, Odisha, India  and the M. Sc.  degree in computer science from Gangadhar Meher University, Odisha, India. She is working toward a Ph.D. degree in computer science as a research scholar with the Dept. of Computer Science, Gangadhar Meher University, Odisha, India. Her research interests include Machine Learning,  Artificial Neural Networks, Pattern Recognition,  Soft Computing, and Bio-informatics. 
\par
$\\$$\\$$\\$$\\$$\\$
\begin{wrapfigure}[12]{l}{35 mm}
	\includegraphics[height=1.5in, width=1.25in]{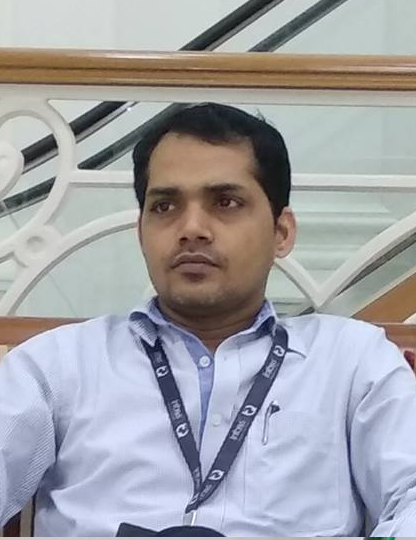}
\end{wrapfigure}\par
\textbf{Ajaya Kumar Tripathy} is a faculty member in the School of Computer Science since 2020. From 2010 to 2019 Dr. Tripathy was working as a faculty member in the department of Computer Science and Engineering, Silicon Institute of Technology, Bhubaneswar, India. Dr. Tripathy holds a Ph.D. degree in computer science from Utkal University in 2015. Dr. Tripathy holds a Professional M.Tech. Degree in e-Government from Trento University, Italy in 2008. In 2007 he completed the M.Tech. Degree in Computer Science from Utkal University, India. From 2008 to 2010 he was a Ph.D. student in the SOA Research Unit at Fondazione Bruno Kessler (FBK) in Trento, Italy. Dr. Tripathy also worked as a research assistant (internee) in Indian Statistical Institute, India, and Createnet Research Center, Trento, Italy. His research interest includes pattern recognition and machine learning, data mining, and soft computing. He has published many research papers in internationally reputed journals and conferences.\par

$\\$

\begin{wrapfigure}[10]{l}{35 mm} 
	\includegraphics[height=1.5in, width=1.25in]{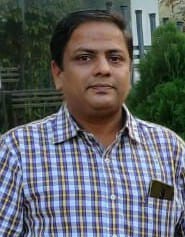}
\end{wrapfigure}\par
\textbf{Laxmi Narayan Padhy} is working as a Professor in the Department of Computer Science and Engineering, Konark Institute of Science and Technology, Bhubaneswar, Odisha, India. He holds a Ph.D. degree in Mathematics from Berhampur University, Berhampur, Odisha, India. In 2007 he completed his MTech degree in Computer Science from Utkal University, India.  His specialization  includes formal language design and compiler for formal languages. His area of research interest includes web services, web service composition, service-oriented architecture, service-oriented architecture monitoring, and applied machine learning. 
\par

$\\$$\\$$\\$$\\$

\begin{wrapfigure}[10]{l}{35 mm}
	\includegraphics[height=1.5in, width=1.25in]{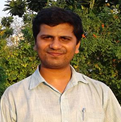}
\end{wrapfigure}\par
\textbf{ Sujita Kumar Kar} is working as an Additional Professor of Psychiatry at King George’s Medical University, Lucknow. His areas of interest include  Neuromodulation in psychiatry, Neuropsychiatry, Suicide prevention, and Culture-bound syndrome. He had more than 370 publications in various national \& international peer-reviewed journals. He is also a peer reviewer of various national and international journals. He is the Editor-in-chief of the Indian Journal of Health, sexuality and culture and Guest Associate Editor in Public Mental Health (Frontiers). He is the editorial board member of the World Journal of Psychiatry, Journal-of-child-and-adolescent-mental-health, and MJDYPV.

$\\$$\\$$\\$$\\$

\begin{wrapfigure}[10]{l}{35 mm} 
	\includegraphics[height=1.5in, width=1.25in]{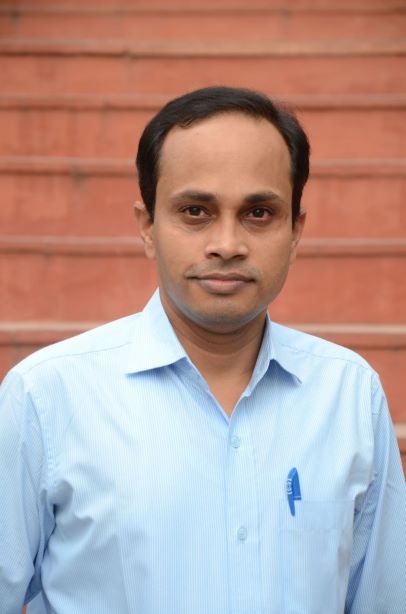}
\end{wrapfigure}\par
\textbf{Susanta Kumar Padhy } is currently working as an Additional Professor in the  Department of Psychiatry, All India Institute of Medical Sciences, Bhubaneswar, India since 2018. Prior to that Dr. Padhy was an Additional Professor in the Department of Psychiatry, PGIMER, Chandigarh, India. Dr. Padhy acquired his MD (Psychiatry)  degree from PGIMER, Chandigarh in 2006. Dr. Padhy was a faculty member in the Department of Psychiatry, PGIMER, Chandigarh, India from 2011 to 2017. Dr. Padhy has published several research articles in the field of Psychiatry of international repute. His research interest includes Severe mental Illness (SMI), Common Mental Disorder (CMD), Child and Adolescent Psychiatry, Community Psychiatry, Mental Health Promotion and prevention, Spirituality, and mental health. 
\par

$\\$

\begin{wrapfigure}[12]{l}{35 mm} 
	\includegraphics[height=1.5in, width=1.25in]{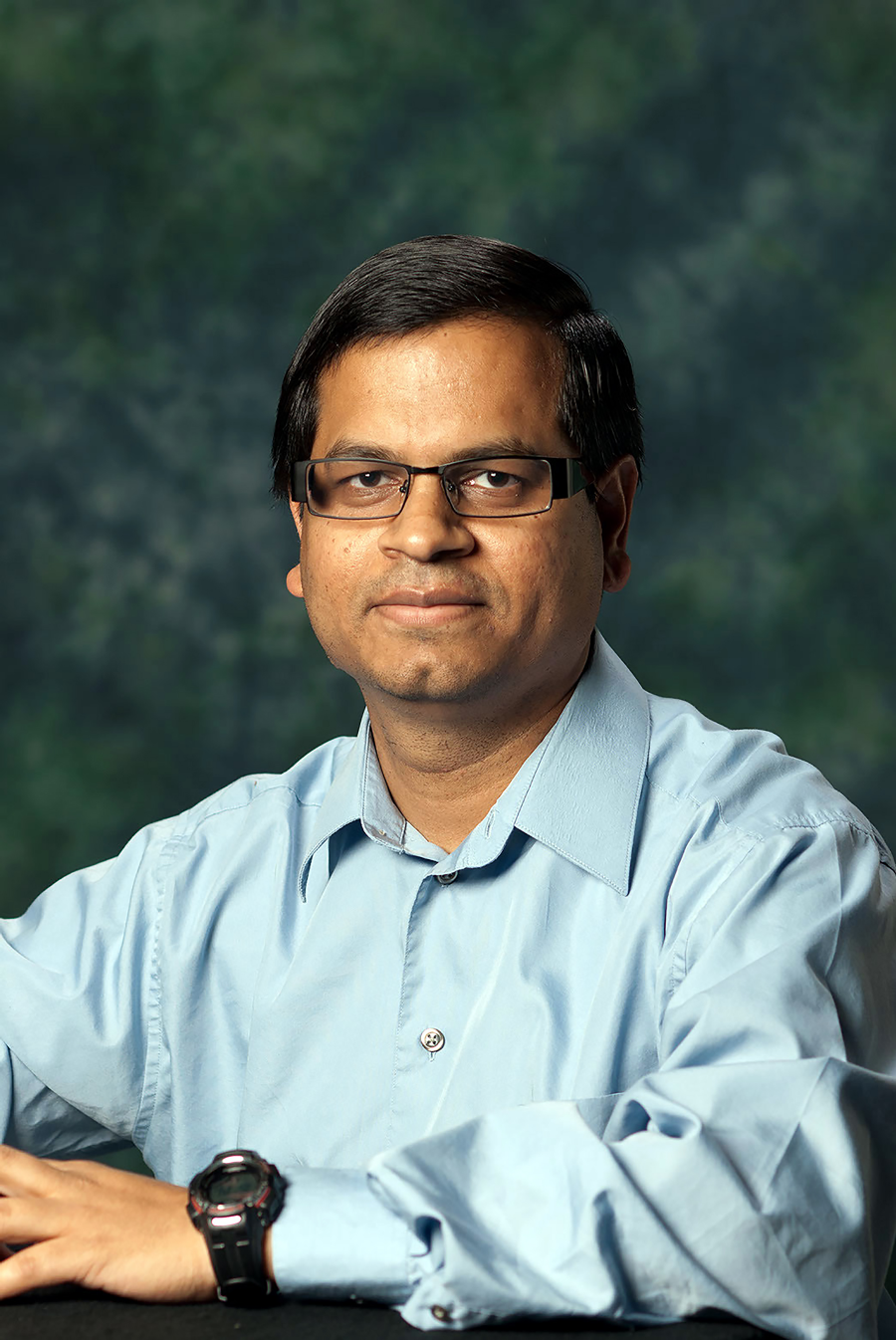}
\end{wrapfigure}\par
\textbf{Saraju P. Mohanty} (Senior Member, IEEE) received the bachelor’s degree (Honors) in electrical engineering from the Orissa University of Agriculture and Technology, Bhubaneswar, in 1995, the master’s degree in Systems Science and Automation from the Indian Institute of Science, Bengaluru, in 1999, and the Ph.D. degree in Computer Science and Engineering from the University of South Florida, Tampa, in 2003. He is a Professor with the University of North Texas. His research is in ``Smart Electronic Systems’’ which has been funded by National Science Foundations (NSF), Semiconductor Research Corporation (SRC), U.S. Air Force, IUSSTF, and Mission Innovation. He has authored 450 research articles, 5 books, and 9 granted and pending patents. His Google Scholar h-index is 47 and i10-index is 203 with 10,300 citations. He is regarded as a visionary researcher on Smart Cities technology in which his research deals with security and energy aware, and AI/ML-integrated smart components. He introduced the Secure Digital Camera (SDC) in 2004 with built-in security features designed using Hardware Assisted Security (HAS) or Security by Design (SbD) principle. He is widely credited as the designer for the first digital watermarking chip in 2004 and first the low-power digital watermarking chip in 2006. He is a recipient of 14 best paper awards, Fulbright Specialist Award in 2020, IEEE Consumer Electronics Society Outstanding Service Award in 2020, the IEEE-CS-TCVLSI Distinguished Leadership Award in 2018, and the PROSE Award for Best Textbook in Physical Sciences and Mathematics category in 2016. He has delivered 18 keynotes and served on 14 panels at various International Conferences. He has been serving on the editorial board of several peer-reviewed international transactions/journals, including IEEE Transactions on Big Data (TBD), IEEE Transactions on Computer-Aided Design of Integrated Circuits and Systems (TCAD), IEEE Transactions on Consumer Electronics (TCE), and ACM Journal on Emerging Technologies in Computing Systems (JETC). He has been the Editor-in-Chief (EiC) of the IEEE Consumer Electronics Magazine (MCE) during 2016-2021. He served as the Chair of Technical Committee on Very Large Scale Integration (TCVLSI), IEEE Computer Society (IEEE-CS) during 2014-2018 and on the Board of Governors of the IEEE Consumer Electronics Society during 2019-2021. He serves on the steering, organizing, and program committees of several international conferences. He is the steering committee chair/vice-chair for the IEEE International Symposium on Smart Electronic Systems (IEEE-iSES), the IEEE-CS Symposium on VLSI (ISVLSI), and the OITS International Conference on Information Technology (OCIT). He has mentored 2 post-doctoral researchers, and supervised 14 Ph.D. dissertations, 26 M.S. theses, and 18 undergraduate projects.
\par

\end{document}